\documentclass{new_tlp}
\usepackage{mathptmx}
\usepackage{amsmath}
\usepackage{hyperref}
\usepackage{graphicx}
\usepackage{subfig}
\usepackage{multirow}

  \title[Online Event Recognition from Moving Vehicles]
        {Online Event Recognition from Moving Vehicles: Application Paper}

  \author[Efthimis Tsilionis et al.]
         {EFTHIMIS TSILIONIS\textsuperscript{1}, NIKOLAOS KOUTROUMANIS\textsuperscript{2}, \authorbreak PANAGIOTIS NIKITOPOULOS\textsuperscript{2}, CHRISTOS DOULKERIDIS\textsuperscript{2} \authorbreak AND ALEXANDER ARTIKIS\textsuperscript{2,1}\authorbreak
         \textsuperscript{1}National Center for Scientific Research `Demokritos', Athens, Greece\authorbreak
         \textsuperscript{2}University of Piraeus, Piraeus, Greece\authorbreak
         (\textit{e-mail:} eftsilio@iit.demokritos.gr, \authorbreak 
         \{koutroumanis, nikp, cdoulk, a.artikis\}@unipi.gr)
         }

\def\rtec{RTEC}

\def\happensAt{\textsf{\small happensAt}}
\def\holdsAt{\textsf{\small holdsAt}}
\def\holdsFor{\textsf{\small holdsFor}}
\def\initiatedAt{\textsf{\small initiatedAt}}
\def\terminatedAt{\textsf{\small terminatedAt}}

\def\true{\textsf{\small true}}

\def\endE{\textsf{\small end}}

\def\prevQ{$q_{i-1}$}

\allowdisplaybreaks

\begin{document}
\nocite{*}% includes all entries of BibTeX database into the list of references.

\maketitle

\begin{abstract}
We present a system for online composite event recognition over streaming positions of commercial vehicles. Our system employs a data enrichment module, augmenting the mobility data with external information, such as weather data and proximity to points of interest. In addition, the composite event recognition module, based on a highly optimised logic programming implementation of the Event Calculus, consumes the enriched data and identifies activities that are beneficial in fleet management applications. We evaluate our system on large, real-world data from commercial vehicles, and illustrate its efficiency. Under consideration for acceptance in TPLP.
\end{abstract}

  \begin{keywords}  
  Event Pattern Matching, Event Calculus, Data Enrichment
  \end{keywords}
  
\section{Introduction}
\label{intro}

The European economy relies to a great extent on commercial vehicle fleets.  According to the European 
Automobile Manufacturers Association\footnote{\url{http://www.acea.be/statistics/article/vehicles-in-use-europe-2017}}, there were over 54 Million commercial vehicles in use in Europe in 2015, and this number is growing every year. Commercial vehicles are equipped with devices emitting information regarding their location and operational status, such as speed and fuel level.
Fleet management applications collect the information emitted from moving vehicles in order to improve the management and planning of transportation services, and enable informed decision-making. Detecting composite events from such data streams can be beneficial for the drivers of commercial vehicles, since they can be informed about their performance, and even prevent dangerous situations. Additionally, the analysis of data generated by such a fleet of vehicles, can help the owners maximize the performance of the fleet.

However, the data produced by a fleet of vehicles is not always sufficient on its own to support advanced vehicle monitoring. External data sources, such as weather information or proximity to points of interest (POIs), can have a significant effect on the movement of the vehicles. For example, fleet management applications can estimate better the fuel consumption of the fleet, by taking into consideration weather information. Furthermore, informing about the presence of locations of interest in a close distance, such as gas stations, can be a significant help both for drivers and fleet operators. Therefore, the integration of positional information with external data sources allows for improved monitoring.

In the context of the Track \& Know project\footnote{\url{https://trackandknowproject.eu/}}, we develop an online fleet management system for the recognition of composite events, that improves the operating efficiency of a commercial fleet. Our system utilizes the GPS (Global Positioning System) traces of moving vehicles along with information emitted by an installed accelerometer device, such as an abrupt acceleration, and information concerning the level of fuel in a vehicle's tank provided by a fuel sensor. These traces are enriched with weather and POI information by a dedicated component for data enrichment. The enriched data are provided as input to a composite event recognition (CER) component, which is based on the `Event Calculus for Run-Time Reasoning' (RTEC). This is a logic programming implementation of the Event Calculus \cite{DBLP:journals/ngc/KowalskiS86} with optimizations for continuous narrative assimilation on data streams \cite{DBLP:journals/tkde/ArtikisSP15,eftsilio_DEBS}.
%
%\rtec\ has proven capable of real-time CER in numerous applications, such as city transport management, public space surveillance from video content \cite{DBLP:journals/tkde/ArtikisSP15} and maritime monitoring \cite{DBLP:journals/geoinformatica/PatroumpasAAVPT17,alevizos2015}. The common case in such streaming environments is for the input events to arrive with variable delays to the CER system. \rtec\ deals with delays by means of windowing; in particular, \rtec\ uses overlapping windows in order to `wait' for delayed events. However, \rtec\ computes every time the complex events (CEs) within a window from scratch, leading to inefficiency. Thus, even if a CE remains unaffected by delays, its intervals will be re-calculated, leading to inefficiency. To overcome this, we present an incremental version of \rtec\ \cite{eftsilio_DEBS}, which updates only the CEs affected by delays, reduces the number of calculations and as a result improves the computational performance.
%
The contributions of this paper are then the following:
\begin{itemize}
\item We provide a high-throughput and scalable solution for the enrichment of mobility data with weather information and nearby POIs.

\item We present a stream reasoning system integrating the component for data enrichment and a logic programming component for recognizing composite events.

\item We illustrate our approach using large, real-world, heterogeneous data streams concerning commercial vehicles. The evaluation validates the robustness and scalability of the system as well as its capacity to operate in real-time.
\end{itemize}

The remainder of this paper is organized as follows. Section \ref{sec:rel} discusses related work, while 
%Section \ref{sys_arch} describes the system architecture. Then, 
Sections \ref{data_enrich} and~\ref{cer} present the main system components. Section \ref{empir} presents our empirical evaluation. Finally, Section \ref{sum} discusses the challenges that we faced during the system development.

\section{Related Work}
\label{sec:rel}
Data enrichment is considered as part of a larger process known as \emph{data integration}, which is a challenging topic, in particular in the context of data sources that provide large volumes of data, often in streaming mode, and in heterogeneous formats~\cite{DBLP:series/synthesis/2015Dong}.  Unfortunately, despite the significance of integrating mobility data with weather, there is a lack of publicly available and reusable systems or tools; our work on weather data integration~\cite{DBLP:conf/edbt/KoutroumanisSGD19} aims to address this limitation. Regarding the enrichment of GPS traces with static locations, also known as points of interest, the problem is essentially known as \emph{distance join}, a variant of spatial joins~\cite{DBLP:journals/tods/JacoxS07}, where records from two data sets are joined if their distance is below a user-specified threshold. Parallel processing of distance join is typically performed in two ways: (a) by repartitioning both data sets to processors in a way that guarantees the correctness of the result, when partitions are processed independently, or (b) by partitioning one data set and broadcast the other to all processors. The latter technique is usually preferred when one of the data sets is relatively small, and we adopt this method here.

Composite event recognition (CER) systems accept as input a stream of time-stamped, `simple, derived events', such as events coming from sensors of moving vehicles, and identify composite events (CE)s of interest --- collections of events that satisfy some pattern. The definition of a CE imposes temporal and, possibly, atemporal constraints on its sub-events (simple, derived events or other CEs).
Numerous CER systems and languages have been proposed in the literature. See \cite{DBLP:journals/csur/CugolaM12,DBLP:journals/csur/AlevizosSAP17,vldbj-cer-survey} for three surveys.
These systems have a common goal, but differ in their architectures, data models, pattern languages and processing mechanisms \cite{DBLP:conf/icdt/GrezRU19}.
For example, many CER systems provide users with a pattern language that is later compiled into some form of automaton (\citeNP{DBLP:conf/cidr/DemersGPRSW07,sasep,DBLP:conf/debs/Schultz-MollerMP09}; Apache {F}link{CEP}\footnote{\url{https://ci.apache.org/projects/flink/flink-docs-stable/dev/libs/cep.html}}). 
The automaton model is used to provide the semantics of the language and/or as an execution framework for pattern matching.
Apart from automata, some CER systems employ tree-based models \cite{DBLP:conf/sigmod/LiuRGGWAM11,DBLP:conf/sigmod/MeiM09}. Again, tree-based formalisms are used for both modeling and recognition, i.e., they may describe the event patterns and the applied recognition algorithm. 

Logic-based approaches to CER have also been attracting considerable attention, since they exhibit a formal, declarative semantics, and at the same time support efficient reasoning \cite{dousson07,DBLP:conf/debs/CugolaM10,DBLP:journals/dss/PaschkeB08}.
We adopt the `Event Calculus for Run-Time reasoning' (RTEC) for our CER engine \cite{DBLP:journals/tkde/ArtikisSP15}, a logic programming implementation of the Event Calculus \cite{DBLP:journals/ngc/KowalskiS86}, that has been used in various application domains, such as maritime monitoring \cite{DBLP:journals/geoinformatica/PatroumpasAAVPT17}.
CE patterns in RTEC are (locally) stratified logic programs. RTEC explicitly represents CE intervals (unlike e.g.~\citeNP{dousson07,DBLP:conf/debs/CugolaM10,lars-rr17}) and thus avoids the related logical problems \cite{paschke05}.
Moreover, and in contrast to state-of-the-art recognition systems, such as the Esper\footnote{\url{http://www.espertech.com/esper/}} engine and SASE \cite{sasep}, RTEC can naturally express hierarchical knowledge by means of well-structured specifications, and consequently employ caching techniques to avoid unnecessary re-computations.

Concerning the Event Calculus literature, a key feature of RTEC is that it includes a windowing technique. No other Event Calculus system (including \citeNP{chittaro96,DBLP:conf/time/CervesatoM00,miller02,DBLP:journals/dss/PaschkeB08,DBLP:journals/igpl/ArtikisS10,DBLP:journals/tist/MontaliMCMA13}) `forgets' or represents concisely the data stream history.

\section{Data Enrichment}
\label{data_enrich}

\begin{figure}[t]
	\centering
	\includegraphics[width=.6\textwidth]{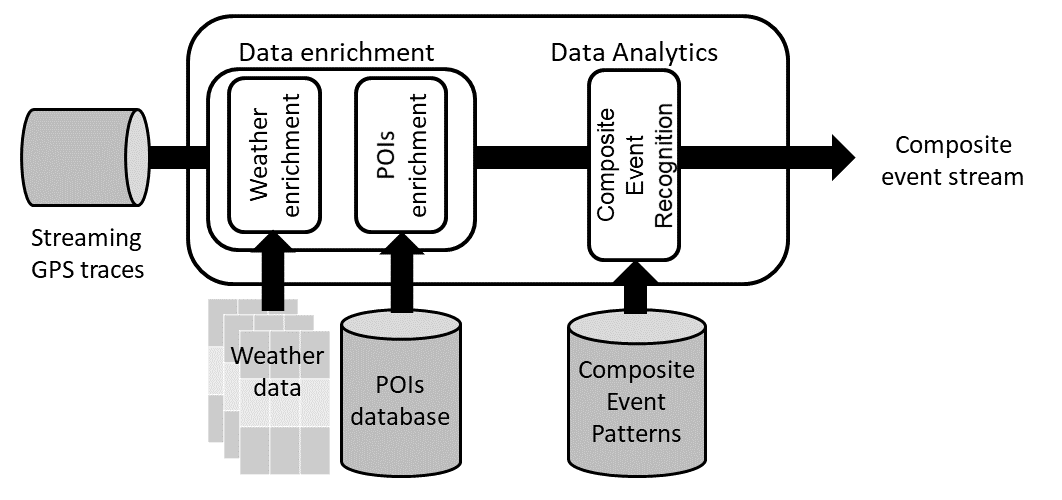}
	\caption{The system architecture for online event recognition from moving vehicles.}
	\label{fig:arch}
\end{figure}

The architecture of our system for online fleet management is depicted in Figure~\ref{fig:arch}. 
The main input is streaming GPS traces from a fleet of moving vehicles, typically provided by a fleet management application. As this streaming data flows in the system, it is enriched with external information, mainly weather data and proximity to points of interest. The enrichment process augments the GPS traces with valuable information, which can be exploited for identifying patterns of composite events (CEs) that would otherwise remain hidden.
Subsequently, a CER module consumes the stream of enriched GPS positions to identify CEs.
%The innovative feature of the proposed architecture is the combination of a data enrichment component with a data analysis component, in order to manifest an online system for complex event recognition, tailored to the requirements of the domain of moving vehicles. 
Moreover, the system architecture is implemented on top of scalable big data frameworks (e.g., Kafka, Spark), thereby exploiting parallelism for data operations either at the level of a cluster of computers or at the level of a single computer (by means of multi-threading).

In the context of this work, data enrichment consists of two modules: weather data enrichment and point of interest (POI) enrichment.
The input is GPS traces from a set of moving vehicles, which contain vehicle id $(v.id)$, position $v.loc=(v.x, v.y)$ and timestamp $(v.t)$, as well as other attributes (e.g., speed, acceleration, etc.). 
The output contains the same set of records, enriched with additional attributes. First, a set of weather attributes, selected according to needs of the application\footnote{In this paper, we are mostly interested in events and their relationship to ice-related attributes.}. Essentially, for each position $v.loc=(v.x, v.y)$ and timestamp $(v.t)$ of a vehicle, we retrieve the values of weather attributes. Second, each position $(v.x, v.y)$ is enriched with a set of POIs $\{p_i\}$ that are located within a user-specified distance threshold $\theta$, i.e., $d(v.loc, p_i) \leq \theta$. 

\subsection{Weather Enrichment}

The weather enrichment module operates in an online manner, by processing the GPS traces record-by-record, as they arrive in the stream.
Internally, its logic is split in two sub-modules; the \emph{Spatio-temporal parser}, which is responsible for extracting the position $(v.loc)$ and timestamp $(v.t)$ from the input record, and the \emph{Weather data obtainer}, which is responsible for the retrieval of weather attribute values associated with the specific spatio-temporal position $(v.loc,v.t)$.

The Spatio-temporal parser, parses each record of the input data and performs some basic data cleaning operations. It checks the spatio-temporal part both for its existence (null or empty values) and validity (valid longitude and latitude values). Both checks are necessary, as GPS traces are typically noisy and may contain errors.
If a value is not valid or missing, then the parser ignores the entire record, and continues with the next one. Each record with valid spatial and temporal information is passed to the Weather data obtainer sub-module, which is responsible of fetching the weather attribute values from the weather data source. 

Weather data is provided as GRIB-formatted files that store gridded meteorological data in binary form. GRIB files are provided by the National Oceanic and Atmospheric Administration (NoAA), which contains data from computer-generated, numerical weather prediction models.
Weather attributes are represented as values on a 2-dimensional (2D) spatial grid divided into cells, where each cell is mapped to a specific geographical area. 
%In some cases, for example for weather attributes whose value changes according to the altitude, the representation is 3-dimensional (3D), also taking different altitude values into account. In this paper, 
We use GRIB files that provide the highest spatial resolution, namely $0.5^\circ\times0.5^\circ$. 
Each day is composed of 4 GRIB files, which are based on the 4 distinct forecast models that run on a daily basis, with times 00:00, 06:00, 12:00, and 18:00. Each GRIB file contains weather attribute values whose validity is for 3 hours after the forecast, i.e., 03:00, 09:00, 15:00, and 21:00, respectively. 

The Weather data obtainer maintains a tree data structure in-memory, organizing the references (paths) of each GRIB file based on their reference time. For example, a file with time 00:00 contains a forecast for 03:00. Given a timestamp $(v.t)$, the tree is searched to locate the nearest GRIB file in terms of its reference time. For instance, given a timestamp 05:10, the forecast at 03:00 is considered as nearest in time, rather than the one at 09:00. 
Then, this file is accessed in order to fetch the value corresponding to the location $(v.loc)$ at hand.
Since there is an overhead when opening a GRIB file, a caching mechanism is used for maintaining handles to open files. 
This is beneficial for sequential requests that are served from the same GRIB file, as repeated open/close operations are avoided, thus saving  processing time. %The underlying structure of the cache mechanism is based on a Red-Black tree data structure, which indexes . Cache is adjustable, as it can handle a specific number of entries simultaneously. The number of enrties is defined as an argument for the initialization of cache.
For a detailed presentation of the architecture of the weather enrichment module refer to~\citeNP{DBLP:conf/edbt/KoutroumanisSGD19}.

\subsection{POIs Enrichment}
\begin{figure}[t]
	\centering
	    \subfloat[POI Enrichment example.]{
		\label{fig:poi_example}
		\includegraphics[width=.24\textwidth]{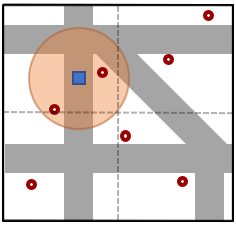}}\qquad
		\subfloat[The POI Enrichment process.]{
		\includegraphics[width=.6\textwidth]{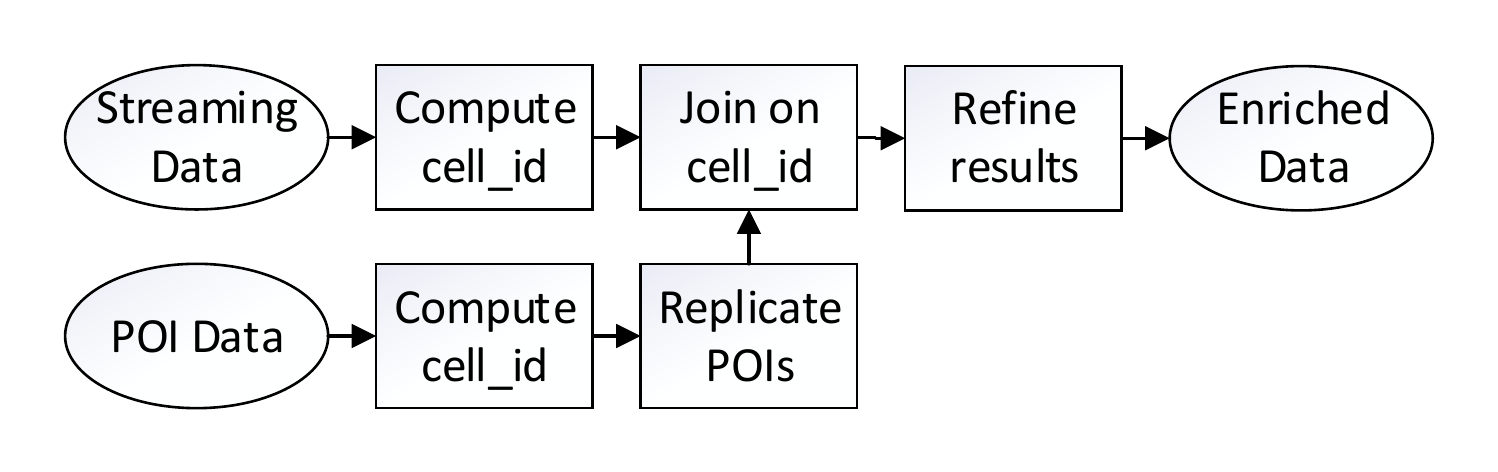}
		\label{fig:poi_enrichment}}%\hspace{1.5cm}
	\caption{POI Enrichment.}
	\label{fig:poi}
\end{figure}

The POI enrichment is implemented as an Apache Spark Structured Streaming job to improve efficiency through parallelized processing. It takes as input (a) the streaming spatio-temporal data set of moving vehicles, (b) a set of POIs containing their spatial information, and (c) a distance threshold $\theta$ expressed in meters. The POI data set, provided by OpenStreetMap, refers to static points of interest described by their spatial location $p.loc=(p.x, p.y)$, name and type of POI. The POI enrichment aims to enrich the spatio-temporal GPS traces of moving vehicles with the information of POIs located at maximum distance $\theta$ from any trace. An example is depicted in Figure~\ref{fig:poi_example}, where the blue rectangle represents a vehicle moving through a city's road network, and the red small circles refer to various places of interest. The circle centered at the vehicle's location with radius $\theta$ encloses all POIs which are located at maximum distance $\theta$ from the vehicle. Hence, our goal is to efficiently identify these nearby POIs and add them to the trace information of the vehicle.

Essentially, the POI enrichment process evaluates a distance join query over the streaming spatio-temporal data set of GPS traces and the static data set of POIs with a maximum distance threshold $\theta$. A naive solution to this problem would join the POI data set with the entire streaming spatio-temporal data set, and then filter out the records that have a joined distance higher than $\theta$. This solution, however, inflicts high computation cost of $O(n \cdot m)$, where $n$ is the number of POIs and $m$ is the number of traces in the streaming spatio-temporal data set, thus reducing the efficiency of our solution.

We propose a more efficient algorithm for computing the distance join query, demonstrated in Figure~\ref{fig:poi_enrichment}. Our premise is to employ a grid that partitions the spatial space into equally sized cells. All records from both data sets can be easily assigned a cell id, based on their associated spatial information. Since the POI data set is static (i.e. does not change while processing the streaming data set), we start by distributing its records to the available computing nodes, based on their corresponding cell ids. We keep the POI records in the nodes’ main memories to enable fast retrieval later. Then, we start processing the spatio-temporal data set, by distributing every streaming record to the corresponding node, based on the computed cell id. That node performs a join operation between the trace record and all the POI records based on their cell id values. The goal is to evaluate the trace's distance join result on a single node, thus reducing the communication complexity of the join operation. To this end, we opt to replicate all POI records to nearby cells, located at maximum distance $\theta$ from the POI. This results to a new POI data set where every cell id is associated with all the POIs located at maximum distance $\theta$ from the cell. Hence, the aforementioned join operation is guaranteed to process all candidate results, without needing any additional communication between the nodes. The last step is to refine the results, by filtering out the records that have a joined distance higher than $\theta$. The computation complexity of this algorithm is significantly reduced to $O(c \cdot m)$, where $c$ is the average number of POIs in a cell, and $m$ is the number of traces in the streaming spatio-temporal data set. In the example of Figure~\ref{fig:poi_example}, the $\theta$ circle spans through three of the cells. The fourth cell (bottom-right) is located at a distance larger than $\theta$ from the location of the vehicle; by pruning the POIs of that cell, our algorithm achieves higher performance without compromising the correctness of the result.

\section{Composite Event Recognition}
\label{cer}

The enriched data stream from moving commercial vehicles is transmitted to the CER module, in order to recognize various types of vehicle activity. All such activities have been formalized in collaboration with the domain experts of the Track \& Know project. Table \ref{tbl:io} presents the input and the output of the CER component, i.e. the Event Calculus for Run-Time reasoning (RTEC). In the following sections we present \rtec\ and illustrate its use for fleet management.

\begin{table}[t]
\caption{Input and Output of CER. The first six input event types accompany the original GPS stream, while the remaining ones are the result of data enrichment. All input events are instantaneous, while all output CEs are durative.}
\vspace{-0.4cm}
\label{tbl:io}
\begin{tabular}{lll}
\hline
\multicolumn{1}{c}{} & \multicolumn{1}{c}{\textbf{Events}} & \multicolumn{1}{c}{\textbf{Description}}\\
\hline
\multirow{8}{*}{\rotatebox[origin=c]{90}{\textbf{Input}}}
&$\mathit{moving(V, S)}$ & Vehicle $V$ is moving with a speed $S$\\
&$\mathit{stopped(V)}$ & Vehicle $V$ is not moving\\
&$\mathit{abruptAcceleration(V)}$ & Vehicle $V$ accelerates abruptly\\
&$\mathit{abruptDeceleration(V)}$ & Vehicle $V$ decelerates abruptly\\
&$\mathit{abruptCornering(V)}$ & Vehicle $V$ turns abruptly\\
&$\mathit{fuelLevel(V, L)}$ & The level of fuel in tank of vehicle $V$ is $L$\\
&$\mathit{iceOnRoad(V)}$ & Vehicle $V$ is moving in an icy road \\
&$\mathit{closeToGas(V)}$ & Vehicle $V$ is near a gas station \\
\hline
\multirow{3}{*}{\rotatebox[origin=c]{90}{\textbf{Output}}}
&$\mathit{highSpeed(V)}$ & Vehicle $V$ exceeds the user-specified speed limit\\
&$\mathit{dangerousDriving(V)}$ & Vehicle $V$ is potentially moving in a dangerous way\\
%$\mathit{unEconomicDriving(V)}$ & Vehicle $V$ consumes fuel inefficiently\\
&$\mathit{reFuelOpportunity(V)}$ & There is refueling opportunity for vehicle $V$\\
\hline
\end{tabular}
\end{table}

\subsection{Run-Time Event Calculus}
\label{rtec}

The time model of \rtec\ is linear and includes integer time-points. If $F$ is a fluent --- a property that is allowed to have different values at different points in time --- the term $F{=}V$ denotes that fluent $F$ has value $V$. \holdsAt$(F{=}V, T )$ is a predicate representing that fluent $F$ has value $V$ at time-point $T$. \holdsFor$(F{=}V, I)$ represents that $I$ is the list of maximal intervals for which $F = V$ holds continuously. \holdsAt\ and \holdsFor\ are defined in such a way that, for any fluent $F$, \holdsAt$(F{=}V, T )$ if and only if $T$ belongs to one of the maximal intervals of $I$ for which \linebreak\holdsFor$(F{=}V, I)$.
An \emph{event description} in \rtec\ comprises rules that express: (a) event occurrences using the \happensAt\ predicate, (b) the effects of events using the \initiatedAt\ and \linebreak\terminatedAt\ predicates, (c) the values of fluents, with the use of the \holdsAt\ and \holdsFor\ predicates, as well as other, possibly atemporal, parameters. Table \ref{tbl:pred} presents the \rtec\ predicates available to the event description developer.

\begin{table}[t]
\caption{Main predicates of \rtec .}
\label{tbl:pred}
\vspace{-0.4cm}
\begin{tabular}{ll}
\hline
\multicolumn{1}{c}{\textbf{Predicate}} & \multicolumn{1}{c}{\textbf{Meaning}}\\
\hline
\happensAt$(E,\enskip T)$ & Event $E$ occurs at time $T$\\
\holdsAt$(F = V,\enskip T)$ & The value of fluent $F$ is $V$ at time $T$\\
\holdsFor$(F=V,\enskip I)$ & $I$ is the list of maximal intervals for which $F=V$ holds continuously\\
\initiatedAt$(F = V,\enskip T)$ & At time $T$ $F=V$ is initiated\\
\terminatedAt$(F = V,\enskip T)$ & At time $T$ $F=V$ is terminated\\
%\footnotesize relative\_ & $I$ is the list of maximal intervals\\
%\footnotesize complement\_ & produced by the relative complement\\
%{\footnotesize all}$(I',L,I)$& of the list of maximal intervals $I'$\\
%& with respect to every list of\\
%& maximal intervals of list $L$\\
%\unionall$(L,I)$ & $I$ is the list of maximal intervals\\
%& produced by the union of the\\
%& lists of maximal intervals of list $L$\\
%\intersectall$(L,I)$ & $I$ is the list of maximal intervals\\
%& produced by the intersection of\\
%& the lists of maximal intervals of list $L$\\
\hline
\end{tabular}
\end{table}

\subsection{Pattern Representation}
\label{rep}

For a fluent $F$, $F=V$ holds at a particular time-point $T$ if $F=V$ has been initiated by an event at some time-point earlier than $T$, and has not been terminated at some other time-point in the meantime. This is an implementation of the law of \emph{inertia}. The time-points at which $F=V$ is initiated (respectively, terminated) are computed with the use of \initiatedAt\ (resp.~\terminatedAt) rules.
$\mathit{highSpeed(V)}$, for example, is a Boolean fluent denoting that a vehicle $V$ is moving with a speed greater than a user-specified threshold $V_{\theta}$:
\begin{equation} \label{pat1}
\begin{aligned}
&\initiatedAt(\mathit{highSpeed(V)} = \true,\enskip T) \gets \\
&\hspace{15pt} \happensAt(\mathit{moving(V, S)},\enskip T), \\
&\hspace{15pt} \mathit{threshold(V, speed, V_{\theta})},\enskip \mathit{S} > V_{\theta}.\\[2pt]
&\terminatedAt(\mathit{highSpeed(V)} = \true,\enskip T) \gets \\
&\hspace{15pt} \happensAt(\mathit{moving(V, S)},\enskip T), \\
&\hspace{15pt} \mathit{threshold(V, speed, V_{\theta})},\enskip \mathit{S} \leq V_{\theta}.\\
%&\terminatedAt(\mathit{highSpeed(V)} = \true, T) \gets \\
%&\hspace{15pt} \happensAt(\mathit{notMoving(V)}, T). \\
&\terminatedAt(\mathit{highSpeed(V)} = \true,\enskip T) \gets \\
&\hspace{15pt} \happensAt(\mathit{stopped(V)},\enskip T).
\end{aligned}
\end{equation}
$\mathit{moving(V, S)}$ and $\mathit{stopped(V)}$ are input events, presented in Table \ref{tbl:io}. $\mathit{threshold}$ is an atemporal predicate recording the numerical thresholds of the patterns --- in this case, the user-specified speed threshold of each vehicle in our knowledge base. Such a predicate supports code transferability, since the use of different thresholds in different applications may be realised by modifying $\mathit{threshold}$ only, and not the pattern specifications. 
Rule-set \eqref{pat1} states that $\mathit{highSpeed(V)} = \true$ is initiated if a $\mathit{moving}$ event is reported for vehicle $V$, and the speed $\mathit{S}$ of $V$ is greater than $V_{\theta}$. Furthermore, $\mathit{highSpeed(V)} = \true$ is terminated if $V$ is moving with a speed less or equal to $V_{\theta}$, or when a $\mathit{stopped}$ event is reported for $V$. By using the \initiatedAt\ and \terminatedAt\ rules of rule-set \eqref{pat1}, \rtec\ computes the maximal intervals $I$ for which $\mathit{highSpeed(V)} = \true$ holds continuously, i.e. \holdsFor$(\mathit{highSpeed(V)} = \true, I)$. This is achieved by first finding all time-points $T_s$ at which $\mathit{highSpeed(V)} = \true$ is initiated, and then, for each $T_s$, retrieving the first time-point $T_f$ after $T_s$ at which $\mathit{highSpeed(V)} = \true$ is terminated. 
Note that, in this formulation of the Event Calculus, \initiatedAt$(F = V, T)$ does not necessarily imply that $F\neq V$ at $T$. (This is similar to the `weak interpretation' of initiation of the Cached Event Calculus, \citeNP{chittaro96}). Similarly, \terminatedAt$(F = V, T)$ does not necessarily imply that $F=V$ at $T$. Suppose that $F=V$ is initiated at time-points 100 and 110 and terminated at time-points 125 and 135 (and at no other time-points). In that case $F=V$ holds at all $T$ such that $100 < T \leq 125$. 

$\mathit{highSpeed(V)}$ is useful indicator on its own, but can also be used to define potentially dangerous driving:
%\rtec\ supports hierarchical definitions, in the sense that one CE may be used to define other, higher-level CEs. Consider e.g. the Boolean fluent $\mathit{dangerousDriving(V)}$ which depends on fluent:
%
\begin{align} \label{pat2}
&\initiatedAt(\mathit{dangerousDriving(V)} = \true,\enskip T) \gets \nonumber \\
&\hspace{15pt} \happensAt(\mathit{abruptAcceleration(V)},\enskip T), \nonumber \\
&\hspace{15pt} \holdsAt(\mathit{highSpeed(V)} = \true,\enskip T).  \nonumber \\
&\initiatedAt(\mathit{dangerousDriving(V)} = \true,\enskip T) \gets \nonumber \\
&\hspace{15pt} \happensAt(\mathit{abruptDeceleration(V)},\enskip T), \nonumber \\
&\hspace{15pt} \holdsAt(\mathit{highSpeed(V)} = \true,\enskip T). \nonumber \\
&\initiatedAt(\mathit{dangerousDriving(V)} = \true,\enskip T) \gets \nonumber\\
&\hspace{15pt} \happensAt(\mathit{abruptCornering(V)},\enskip T), \\
&\hspace{15pt} \holdsAt(\mathit{highSpeed(V)} = \true,\enskip T). \nonumber  \\
&\initiatedAt(\mathit{dangerousDriving(V)} = \true,\enskip T) \gets \nonumber \\
&\hspace{15pt} \happensAt(\mathit{iceOnRoad(V)},\enskip T), \nonumber \\
&\hspace{15pt} \holdsAt(\mathit{highSpeed(V)} = \true,\enskip T). \nonumber \\[2pt]
%&\terminatedAt(\mathit{dangerousDriving(V)} = \true,\enskip T) \gets \nonumber \\
%&\hspace{15pt} \happensAt(\mathit{notMoving(V)},\enskip T). \nonumber \\
&\terminatedAt(\mathit{dangerousDriving(V)} = \true,\enskip T) \gets \nonumber \\
&\hspace{15pt} \happensAt(\endE\mathit{(highSpeed(V)} = \true),\enskip T). \nonumber \\
&\terminatedAt(\mathit{dangerousDriving(V)} = \true,\enskip T) \gets \nonumber \\
&\hspace{15pt} \happensAt(\mathit{stopped(V)},\enskip T). \nonumber
\end{align}
$\mathit{abruptAcceleration}$, $\mathit{abruptDeceleration}$ and $\mathit{abruptCornering}$ are instantaneous input events provided by the  accelerometer device installed in each commercial vehicle (see Table \ref{tbl:io}). \linebreak$\mathit{iceOnRoad(V)}$ is a weather event emitted by the data enrichment module and states that in the location of $V$ the road is slippery due to ice.
\endE$(F{=}V)$ is a built-in \rtec\ event indicating the ending points of each maximal interval for which $F=V$ holds continuously. 
According to rule-set \eqref{pat2}, therefore, $\mathit{dangerousDriving(V)} = \true$ is initiated when a vehicle $V$ is engaged in a harsh driving event, such as abrupt acceleration, breaking or cornering, or when there is ice on the road and $V$ has speed above the user-specified threshold. % In all cases, we demand the time-points each one of the four events is taking place to be included in the maximal intervals during which $V$ is moving with a speed that exceeds the user-specified threshold. This is accomplished with the use of the \holdsAt\ predicate.
$\mathit{dangerousDriving(V)} = \true$ is terminated when the speed of vehicle $V$ goes below the user-specified threshold, or when it stops moving.
$\mathit{dangerousDriving(V)}$ is thus useful for driver behavior analysis and safety.

Companies owning commercial fleets place emphasis on fuel consumption. One way to achieve this, is detecting opportunities for refueling. Consider the formalisation below:
\begin{equation} \label{pat4}
\begin{aligned}
&\initiatedAt(\mathit{reFuelOpportunity(V)} = \true,\enskip T) \gets \\
&\hspace{15pt} \happensAt(\mathit{closeToGas(V)},\enskip T), \\
&\hspace{15pt} \holdsAt(\mathit{highSpeed(V)} = \true,\enskip T), \\
&\hspace{15pt} \happensAt(\mathit{fuelLevel(V, L)},\enskip T), \\
&\hspace{15pt} \mathit{threshold(V, fuel, V_{tank})},\enskip \mathit{L} < \frac{V_{tank}}{2}.\\[2pt]
&\terminatedAt(\mathit{reFuelOpportunity(V)} = \true,\enskip T) \gets \\
&\hspace{15pt} \happensAt(\mathit{fuelLevel(V, L)},\enskip T), \\
%&\hspace{15pt} \happensAt(\mathit{closeToGas(V)}, T), \\
&\hspace{15pt} \mathit{threshold(V, fuel, V_{tank})},\enskip \mathit{L} \geq \frac{V_{tank}}{2}. \\
%&\hspace{15pt} \happensAt(\endE\mathit{(notEconomicDriving(V)} = \true),\enskip T).
\end{aligned}
\end{equation}
$\mathit{closeToGas(V)}$ is a spatial relation computed by the data enrichment module (see Table \ref{tbl:io}), indicating that a vehicle $V$ is close to a gas station, which is a type of point of interest.
$\mathit{fuelLevel(V, L)}$ is an instantaneous input event emitted by the fuel sensor of each vehicle. $\mathit{threshold(V, fuel, V_{tank})}$ records  the tank size of vehicles. 
According to rule-set \eqref{pat4}, our system starts flagging that vehicle $V$ should refuel when it is close to a gas station, its speed is above the user-specified threshold (implying uneconomic driving), and the fuel level is lower than half of the tank size.
Moreover, we stop flagging the need to refuel when the fuel is more than half of the tank size.  

\section{Implementation and Empirical Analysis}
\label{empir}

\begin{figure}[t]
\centering
\includegraphics[width=0.4\textwidth]{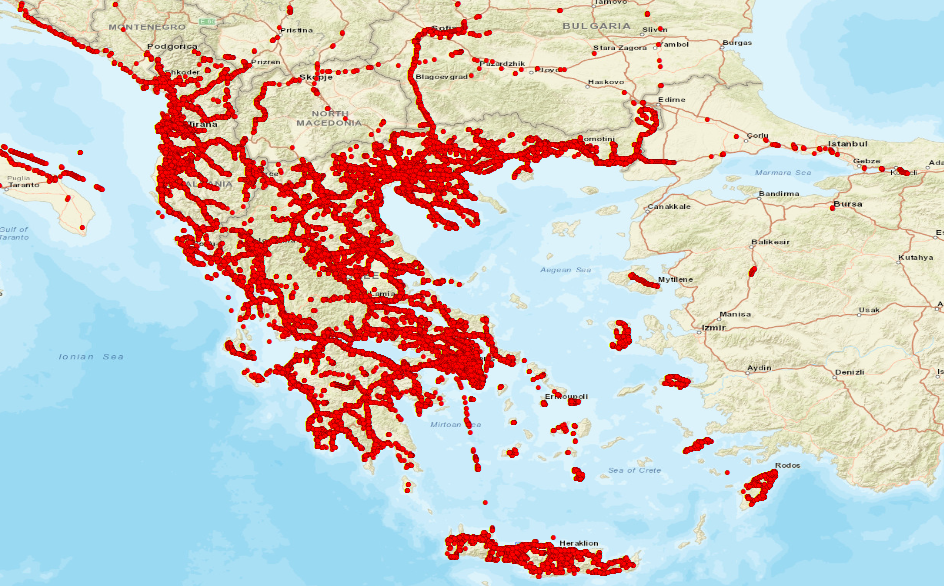}
\caption{Vehicle position signals of the data set.}
\label{fig:geo}
\end{figure}

\subsection{Experimental Setup}
\label{empir:exp_set}

We evaluated our system on real-world positional data of vehicles, provided by Vodafone Innovus\footnote{\url{https://www.vodafoneinnovus.com/}}, our partner in the Track \& Know project, which offers fleet management services. Figure~\ref{fig:geo} illustrates the geographical coverage of the data, practically covering Greece and some surrounding countries, for a temporal duration of 1 month.  
The fleet data contains approximately 4M records and is 527 MB in the form of CSV files. We replayed these records, according to their timestamps, in order to simulate a streaming environment. The records are enriched with weather information, acquired by 120 GRIB files with total size 7.4 GB. The POIs were retrieved from OpenStreetMap; we selected only the POIs referring to gas stations which resulted to approximately 140K POIs. %with size 8 MBs.
%The dataset used for the experiments of the CER module ($\approx 44.2M$ records) concerns 364 vehicles, where each record is enriched with weather data and proximity to POIs. 
The data enrichment module operated on a VM running the CentOs 7.6.1810 operating system, on a hardware with Intel Xeon Processor and 4GB RAM. 
The CER module operated on a computer with 8 cores (Intel(R) Core(TM) i7-7700 CPU @ 3.6GHz) and 16 GB of RAM, running Ubuntu 16.04 LTS 64-bit and YAP Prolog 6.2.2.

\subsection{Data Enrichment}
\label{empir:den}

\begin{figure}[t]
\centering
\subfloat[Varying the number of CPU cores.]{
\label{fig:exp_enrich_cores}
\includegraphics[width=0.35\textwidth]{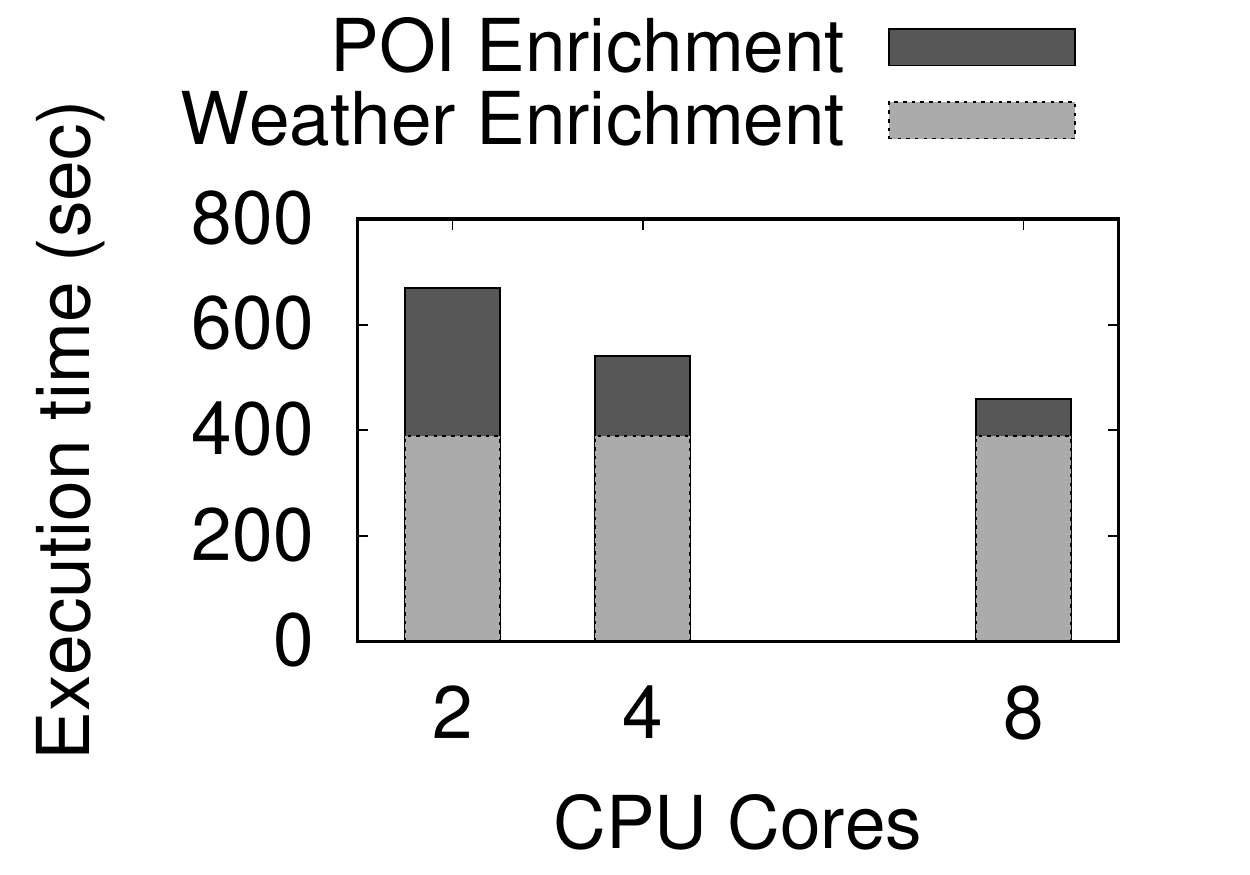}
}
\subfloat[Varying the distance factor $\theta$.]{ 
\includegraphics[width=0.35\textwidth]{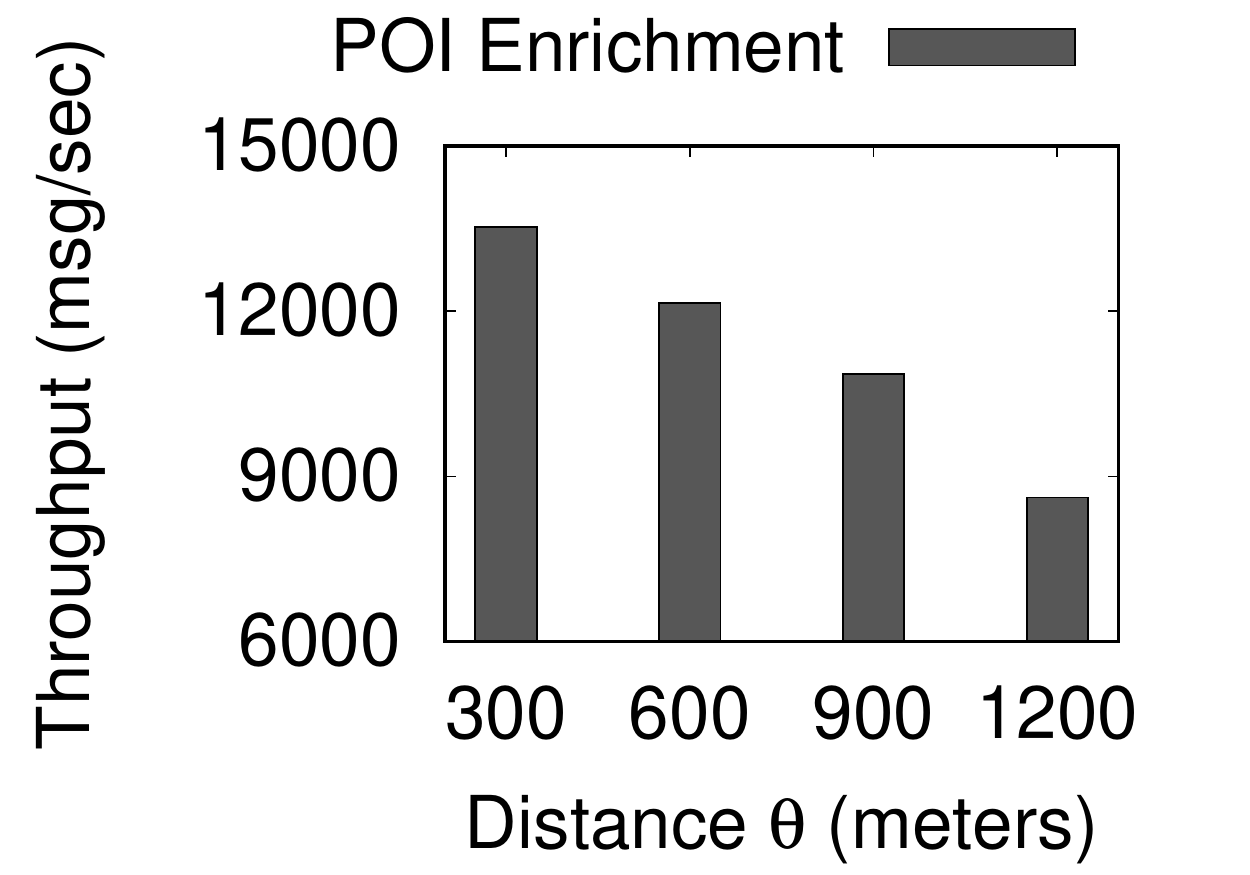}
\label{fig:exp_enrich_theta}
}
\caption{Performance of enrichment process.}
\label{fig:exp_enrich}
\end{figure}

The vehicle position signals were loaded in an Apache Kafka topic, consisting of 6 partitions sorted by the date field.
The topic was consumed by the weather enrichment component, and the poll timeout was set to 1 sec\footnote{This is the time that defines a batch of messages fetched for processing.}. 
%The input records are enriched with 24 weather attributes. 
%The data enrichment process took place on the Apache Kafka streaming platform. Specifically, the sample data was loaded on a Apache Kafka topic, consisted of 6 partitions, in which the records were sorted by date field. 
The POI enrichment runs as a separate job, and consumes weather-enriched data that were output in an intermediate Kafka topic, and was configured to use 2 GB of RAM. 
In this set of experiments, we report on the performance of data enrichment, using total execution time and throughput as main metrics. 

%execution time needed to enrich the data for both the enrichment components, as well as the effect of the $\theta$ parameter on the throughput of the POI enrichment. 
%Since the POI enrichment component benefits by parallelization through multithreading, we vary the number of CPU cores provided to the POI component, to evaluate its efficiency. The weather enrichment component uses a single CPU core, thus we report constant execution time for this set of experiments.

Figure~\ref{fig:exp_enrich} depicts the performance results of data enrichment. Figure~\ref{fig:exp_enrich_cores} reports the total execution time for data enrichment when increasing the number of CPU cores. 
The weather enrichment was completed in 389 sec, corresponding to throughput values of 9,792 messages/sec. Notice that this value is constant in the figure, since weather enrichment does not use parallelization.
The cache of the weather data obtainer reached 99.96\% hit ratio. This  high ratio was expected, since the records are temporally sorted. 
%As demonstrated in Figure~\ref{fig:exp_enrich_cores}, the enrichment procedure was completed in 389 sec with throughput 9792 messages/sec.
The POI enrichment results correspond to distance threshold $\theta=300$m. 
As shown in the figure, the total execution time drops when increasing the parallelism from 2 to 8 CPU cores. This result shows that the POI enrichment component can exploit the availability of more CPUs and scale  based on the available resources. 

%During the first set of experiments (Figure~\ref{fig:exp_enrich_cores}) we configuted the $\theta$ setting to 300 meters. The execution time of the POI enrichment component was measured to 281, 152 and 70 seconds for 2, 4 and 8 CPU cores respectively. This means, that the throughput is roughly 6600 messages per CPU core per second. The POI component performs better when using more CPUs, exploiting the scalabilty of the available resources. 

Figure~\ref{fig:exp_enrich_theta} shows the throughput of POI enrichment obtained from increasing the value of $\theta$, while fixing the number of cores to 2. Higher values of $\theta$ result in having more POIs associated with positions of vehicles, as $\theta$ practically defines what is considered as proximity. Also, higher values of $\theta$ imply that a larger spatial area around each vehicle's position needs to be examined, leading to decreased performance. 
However, when increasing  $\theta$ by a factor of 4, the performance is decreased less than 50\%. Hence, our POI enrichment component is efficient for even higher $\theta$ values. Also, it can achieve even better performance by exploiting more CPU resources, as already shown in Figure~\ref{fig:exp_enrich_cores}.

% Input from UniPi

\subsection{Composite Event Recognition}
\label{empir:cer}

In \rtec , the CER process involves the computation of the maximal intervals of fluents. This process takes place at specified query-times $q_1, q_2, \dots$\ . CER at each query-time $q_i$ is performed over the input events that fall within a specified interval, the `working memory' or window $\omega$. All input events outside the window are discarded and not considered during recognition. This means that at each query-time $q_i$, CER depends only on the events that took place in the interval $(q_i-\omega ,q_i]$. The size of $\omega$ as well as the temporal distance between two consecutive query-times --- the slide step $(q_i -$\prevQ$)$ --- are user-specified.
\begin{figure}[t]
\centering
\includegraphics[width=0.6\textwidth]{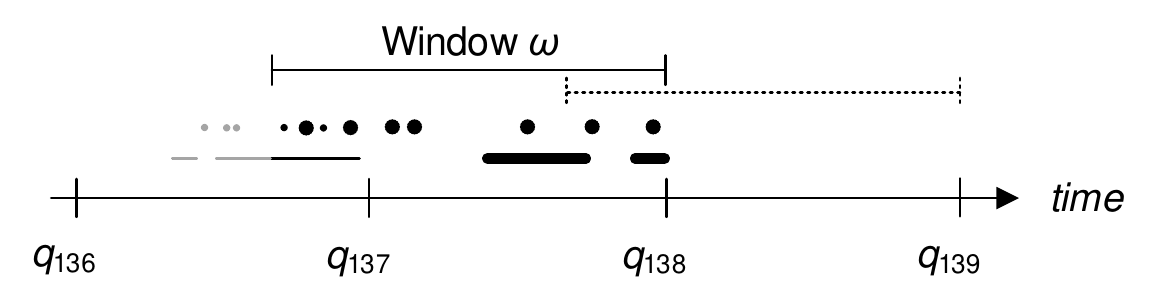}
\caption{Composite event recognition in \rtec.} 
\label{rtec:fig}
\end{figure}
Figure \ref{rtec:fig} illustrates the recognition process of \rtec . Occurrences of instantaneous input events are displayed as dots and those of durative input events as line segments. For CER at query-time  $q_{138}$, only the events marked in black are considered, whereas the greyed out ones are neglected. Assume that the events marked in bold arrive after $q_{137}$. Therefore, two input events are delayed and by using a window size larger than the slide step, these two events are not lost and considered at $q_{138}$. In the analysis that follows, we restrict attention to overlapping windows, i.e. windows longer than the slide step.

At each query-time $q_i$, \rtec\ computes from scratch the intervals of CEs, without considering the computations of previous windows. In the case of significant delays in the input stream, this simple approach is the best option. However, in cases where CEs are unaffected by delays, computing their intervals from scratch is redundant. To address this issue, we recently developed a process for computing incrementally the maximal intervals of a CE \cite{eftsilio_DEBS}. Consider the first initiation rule of rule-set \eqref{pat2} again, and assume that at query-time $q_i$ a delayed arrival of $\mathit{abruptAcceleration(V)}$ arrived to the CER system or/and a new interval was computed for $\mathit{highSpeed(V)}$. In both cases, a new initiation may have to be computed for $\mathit{dangerousDriving}$. To calculate new initiation points, we use the following \textit{delta} rules (the remaining initiation rules and the termination rules of rule-set \eqref{pat2} are handled similarly):
\begin{equation} \label{rdelta}
\begin{array}{ll}
\begin{aligned}
&\initiatedAt(\mathit{dangerousDriving(V)} = \true, \enskip T) \gets \\
&\hspace{15pt} \Big[\happensAt(\mathit{abruptAcceleration(V)},\enskip T)\Big]^{ins}  , \\
&\hspace{15pt} \Big[\holdsAt(\mathit{highSpeed(V)} = \true,\enskip T)\Big]^{\textsf{Q}_{\textsf{i}}}  .
\end{aligned} & \textrm{(a)} \\ \\
\begin{aligned}
&\initiatedAt(\mathit{dangerousDriving(V)} = \true, \enskip T) \gets \\
&\hspace{15pt} \Big[\happensAt(\mathit{abruptAcceleration(V)},\enskip T)\Big]^{\textsf{Q}_{\textsf{i}} \setminus ins} , \\
&\hspace{15pt} \Big[\holdsAt(\mathit{highSpeed(V)} = \true,\enskip T)\Big]^{ins}  .
\end{aligned} & \textrm{(b)}
\end{array}
\end{equation}
The superscripts of these rules express the evaluation set of the time argument $T$. In rule \eqref{rdelta}(a), $\mathit{abruptAcceleration(V)}$ is evaluated only over the occurrences that arrived to the CER system between $q_{i-1}$ and $q_{i}$, i.e.~the occurrences in set $ins$. The time-points in  $ins$ are examined against all the intervals of $\mathit{highSpeed(V)} = \true$ overlapping the current window (set $\textsf{Q}_{\textsf{i}}$). Rule \eqref{rdelta}(b) is similar to \eqref{rdelta}(a), but has a small modification which ensures that derivations are not repeated. In this rule, only the intervals computed at $q_i$ are considered for $\mathit{highSpeed(V)} = \true$ (set $ins$). For $\mathit{abruptAcceleration(V)}$ the occurrences within the current window that arrived by the previous query-time $q_{i-1}$ (set $\textsf{Q}_{\textsf{i}} \setminus ins$) are used.
\begin{figure}[t]
\centering
\subfloat[Average recognition time.]{ 
\includegraphics[width=0.35\textwidth]{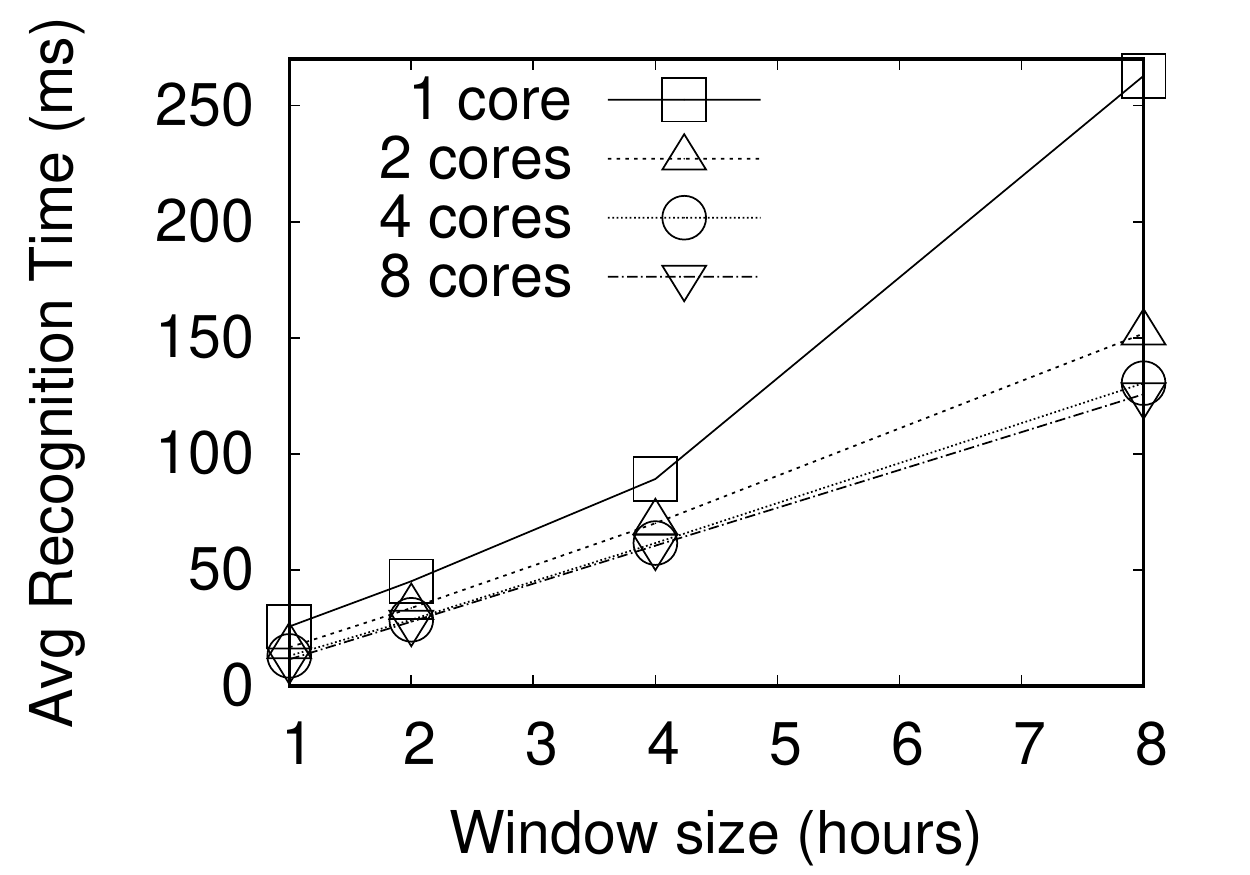}
}
\subfloat[Throughput.]{ 
\includegraphics[width=0.35\textwidth]{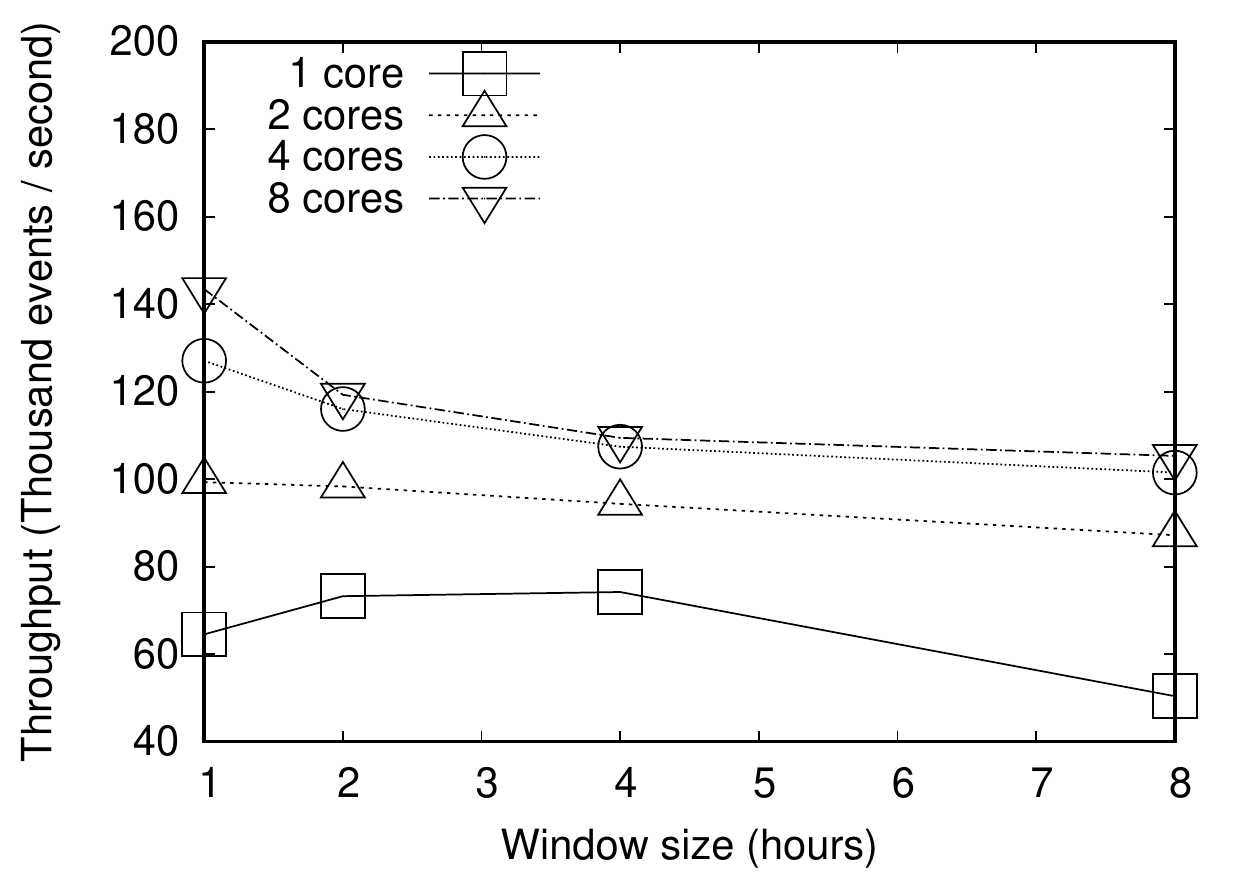}
}
\caption{CER under varying window sizes and parallel configurations.}
\label{fig:rtec_res}
\end{figure}

We performed two sets of experiments. First, since the data set is temporally sorted, we evaluated the performance of \rtec\ without incremental reasoning, on varying window sizes and parallel executions. Second, we injected artificial delays to the data set, to simulate online processing, and thus compared \rtec\ with and without incremental reasoning. Figure \ref{fig:rtec_res} shows the results of the first set of experiments. Initially, we used a single processor to perform CER. Then, we run \rtec\ in parallel, by launching different instances of the engine, each one operating on a different processing core. Each \rtec\ instance performed CER for a different set of vehicles. For example, in the case of four processing cores each \rtec\ instance was responsible for one quarter of operating vehicles. In all sets of experiments the input was the same, that is, there was no data distribution.

We varied the window size $\omega$ from 1 to 8 hours and the slide step was always equal to the size of the window. In the absence of delays, it is redundant to have overlapping windows. Figure \ref{fig:rtec_res}(a) presents the recognition times of \rtec\ in CPU milliseconds (ms), while Figure \ref{fig:rtec_res}(b) presents the throughput. The empirical analysis shows that \rtec\ is capable of real-time CER even when operating on a single core. Additionally, running \rtec\ in parallel leads to significant performance gains even without data distribution.

The second set of experiments concerns out-of-order streams where we  compared the performance of \rtec\ and its incremental extension. We injected artificially delays into the data set. We performed three experiments, each time varying the amount of input events being delayed. We selected uniformly 5\%, 10\% and 20\% of the total events to be delayed. We used a uniform distribution for selecting events, since we assume that each event has the same probability to be delayed. In order to mimic reality as much as possible, we used a Gamma distribution to choose the extent of delay. (The Gamma distribution has a shape parameter $k = 2$ and a scale parameter $\theta = 2$.) Thus, a delay small in time has a higher probability to be imposed in a selected event. The average delay time, in all settings, is approximately 8 hours.
\begin{figure}[Ht]
\centering
\subfloat[5\% delayed events]{ 
\includegraphics[width=0.3\textwidth]{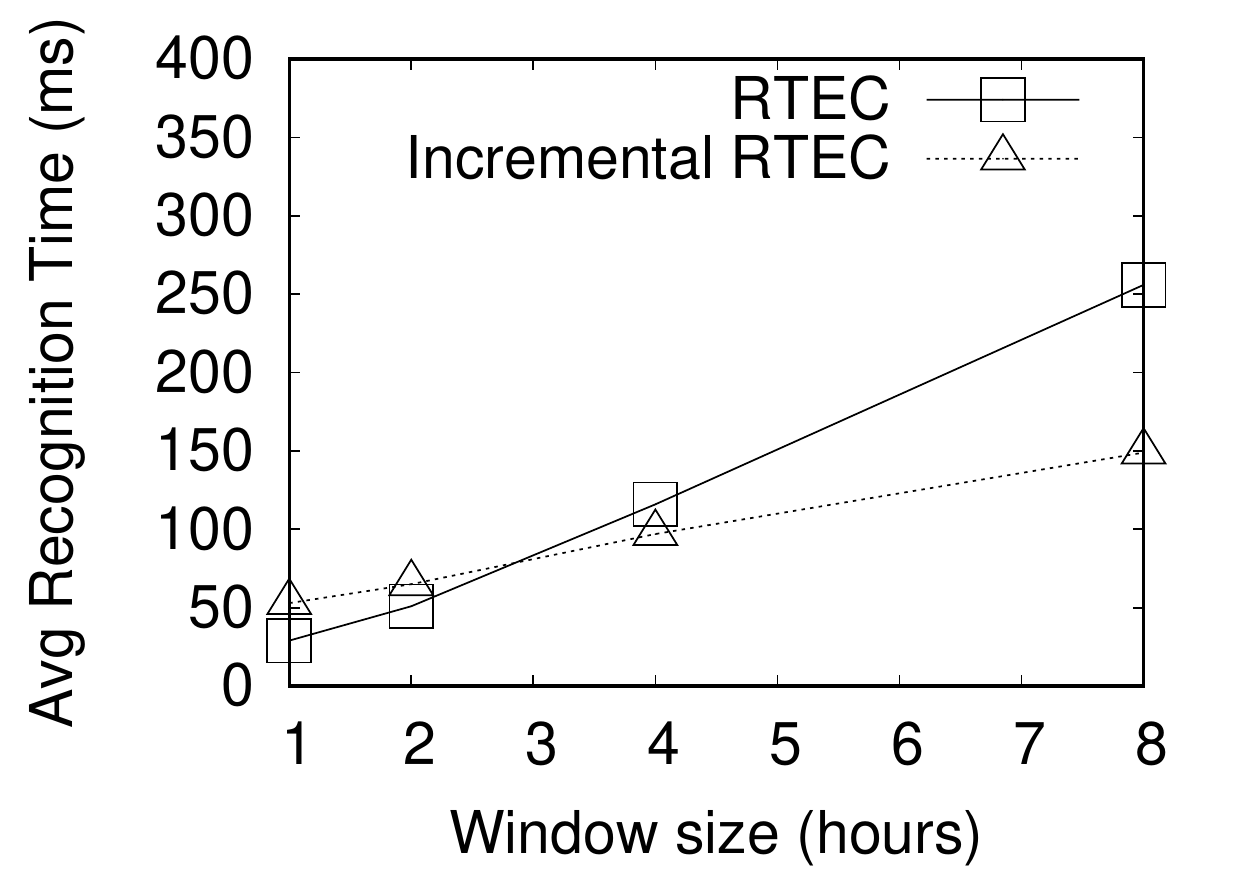}
}
\subfloat[10\% delayed events]{ 
\includegraphics[width=0.3\textwidth]{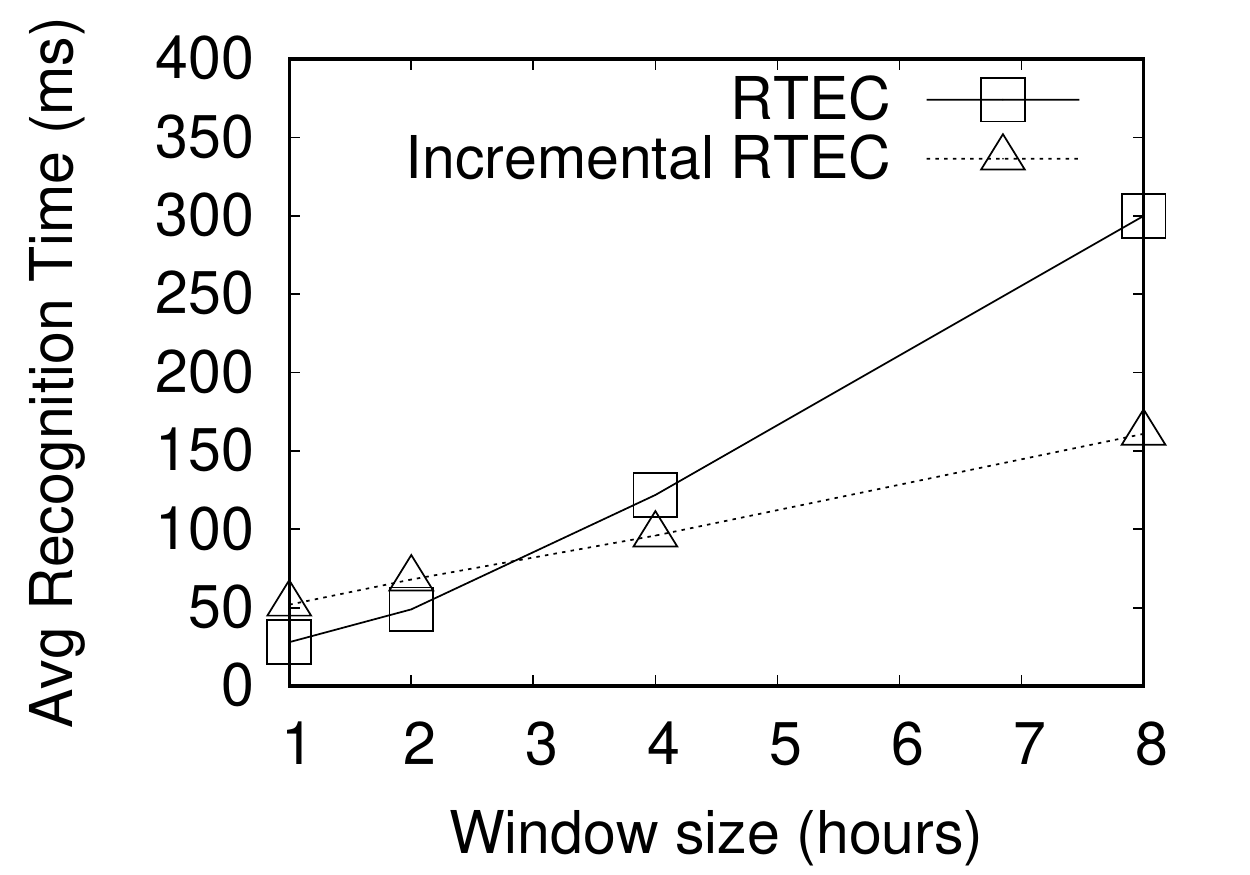}
}
\subfloat[20\% delayed events]{ 
\includegraphics[width=0.3\textwidth]{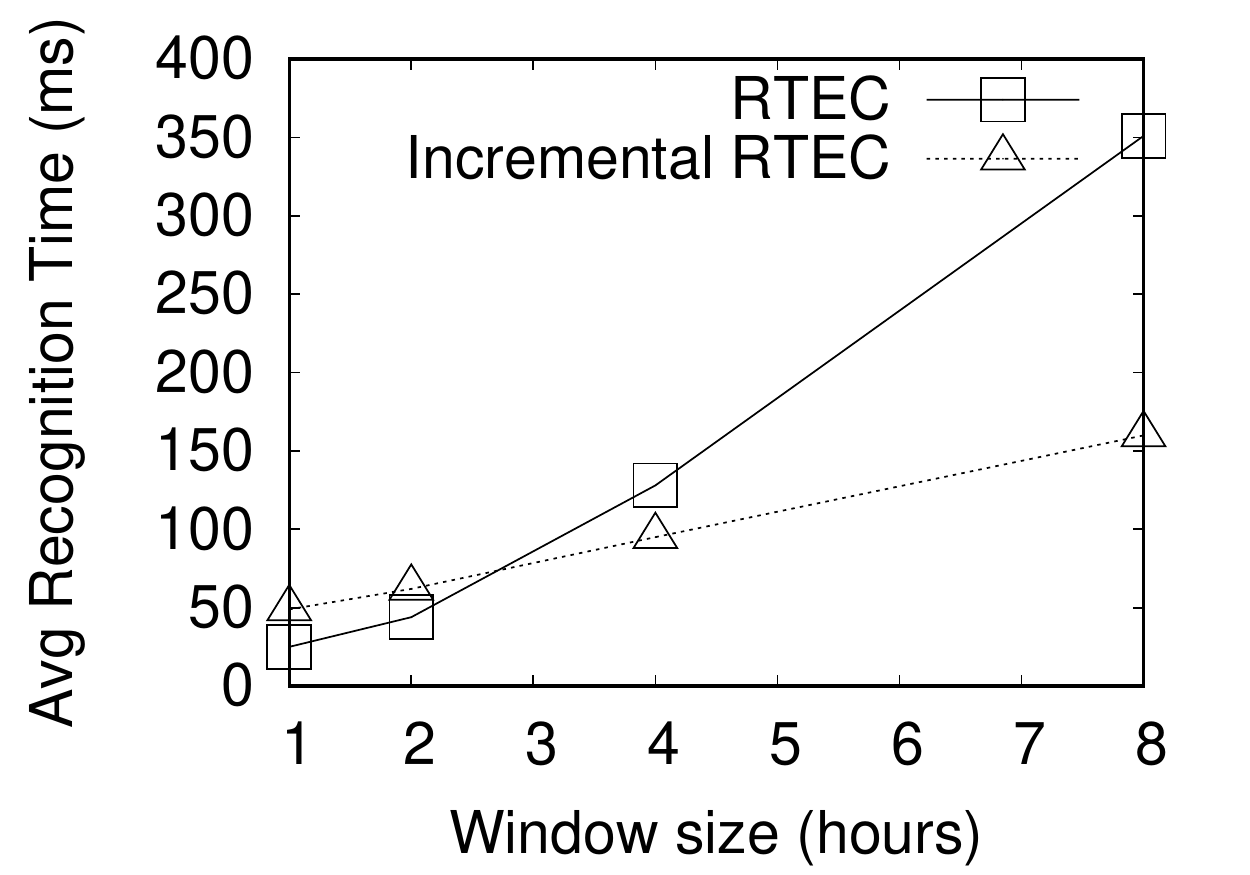}
}
\caption{Effects of incremental reasoning.}
\label{fig:comp}
\end{figure}

Figures \ref{fig:comp}(a-c), display the average recognition times in CPU milliseconds for windows ranging from 1 hour to 8 hours, and a slide step of 1 hour. As shown in Figures \ref{fig:comp}(a-c), the incremental version of \rtec\ outperforms the non-incremental one in the largest windows, i.e. those of 4 and 8 hours. In other words, the performance improvement becomes more profound as the overlap between consecutive windows increases.

\section{Discussion}
\label{sum}

We presented a stream reasoning system for online fleet management. % The proposed system integrates a novel component for CE recognition which augments the original GPS traces of moving vehicles with additional information, such as weather data and proximity to POIs. The CE definitions allowed us to perform a realistic evaluation in terms of performance. Our experimental evaluation showed that our system can support real-time decision-making in the fleet management domain using real data.
We opted for a separation of activities among different modules. %Consider that the fleet management requires a combination of temporal reasoning with spatial reasoning. 
Delegating data enrichment to a separate module allows for the effective integration of spatial reasoning with temporal reasoning for online CER. Additionally, the use of a dedicated module for data enrichment allows to combine heterogeneous data sources in an efficient way.
The empirical evaluation on real-world data illustrated the scalability of our system as well as its capacity to operate in real-time.

A challenge that we faced in the development and deployment of our system is the memory leak of various Prolog implementations, such as YAP and SWI-Prolog, on continuous queries. To address this issue, we sometimes had to store the recognised CEs in order to restart the engine, which is suboptimal in online processing.

Similar to other Big Data projects \cite{DBLP:journals/tkde/ArtikisSP15,DBLP:journals/geoinformatica/PatroumpasAAVPT17,DBLP:conf/bigdataconf/ArtikisWGKG13}, the datasets of the Track \& Know project did not come with a ground truth of CEs. 
One way to address this issue is to construct the CE patterns in close collaboration with domain experts. This is what we did in Track \& Know. 
However, although the domain experts of the project have some idea about the CEs of interest, the precise conditions in which a CE should be recognized are not always clear. The use of \rtec\ facilitated the interaction of CE pattern developers and domain experts. Patterns in the language of \rtec\ were understood, and sometimes directly modified by the domain experts. To facilitate this process further, we have been developing a simple language for \rtec , with the aim of supporting people who are not familiar with the Event Calculus or (logic) programming \cite{DBLP:conf/commonsense/VlassopoulosA17}. A compiler translates, in a process transparent to the user, a specification in the simple language to an \rtec\ event description that may be subsequently used for continuous query computation.

To allow for accuracy evaluation, we implemented a way of visualising our recognised CEs by means of videos\footnote{See \url{http://cer.iit.demokritos.gr/}}. Our aim is to enable domain experts offer feedback on our recognised events, i.e.~classify them as true or false positives. Using such videos, domain experts were able to perform a preliminary accuracy assessment. The findings of this assessment indicated that some CE intervals ended later than anticipated. This is due to the fact that the position signals of vehicles can be sparse. For example, there are some extreme cases in which there are 24 hours between two consecutive positional signals of the same vehicle, most likely indicating different trips. In some of these cases, the CE was terminated on the signal of the subsequent trip, i.e. the termination was delayed.  In order to deal with this issue, we can directly use the `deadlines' mechanism of RTEC, according to which a CE is automatically terminated after a designated number of time-points since the last initiation. A systematic accuracy evaluation based on expert feedback, using the aforementioned visualisations, is part of our current work. 

We also aim to refine the manually constructed CE patterns by means of a recently developed technique for semi-supervised learning  \cite{DBLP:journals/ml/MichelioudakisA19}. The input of this technique will be the expert feedback as described above, as well as a small set of labels that may be provided with minimal resources by domain experts.

\section*{Acknowledgments}

This work was funded by European Union's Horizon 2020 research and innovation programme Track \& Know ``Big Data for Mobility Tracking Knowledge Extraction in Urban Areas'', under grant agreement No 780754.


%\bibliographystyle{ACM-Reference-Format}
%\bibliography{ICLP19}

\begin{thebibliography}{29}

%%% the Web address.

\ifx \showCODEN    \undefined \def \showCODEN     #1{\unskip}     \fi
\ifx \showDOI      \undefined \def \showDOI       #1{#1}\fi
\ifx \showISBNx    \undefined \def \showISBNx     #1{\unskip}     \fi
\ifx \showISBNxiii \undefined \def \showISBNxiii  #1{\unskip}     \fi
\ifx \showISSN     \undefined \def \showISSN      #1{\unskip}     \fi
\ifx \showLCCN     \undefined \def \showLCCN      #1{\unskip}     \fi
\ifx \shownote     \undefined \def \shownote      #1{#1}          \fi
\ifx \showarticletitle \undefined \def \showarticletitle #1{#1}   \fi
\ifx \showURL      \undefined \def \showURL       {\relax}        \fi
% invisible to TeX
\providecommand\bibfield[2]{#2}
\providecommand\bibinfo[2]{#2}
\providecommand\natexlab[1]{#1}
\providecommand\showeprint[2][]{arXiv:#2}

\bibitem[\protect\citeauthoryear{Alevizos, Skarlatidis, Artikis, and
  Paliouras}{Alevizos et~al\mbox{.}}{2017}]%
        {DBLP:journals/csur/AlevizosSAP17}
\bibfield{author}{\bibinfo{person}{Elias Alevizos}, \bibinfo{person}{Anastasios
  Skarlatidis}, \bibinfo{person}{Alexander Artikis}, {and}
  \bibinfo{person}{Georgios Paliouras}.} \bibinfo{year}{2017}\natexlab{}.
\newblock \showarticletitle{Probabilistic Complex Event Recognition: {A}
  Survey}.
\newblock \bibinfo{journal}{\emph{{ACM} Comput. Surv.}} \bibinfo{volume}{50},
  \bibinfo{number}{5} (\bibinfo{year}{2017}), \bibinfo{pages}{71:1--71:31}.
\newblock
\urldef\tempurl%
\url{https://doi.org/10.1145/3117809}
\showDOI{\tempurl}


\bibitem[\protect\citeauthoryear{Artikis and Sergot}{Artikis and
  Sergot}{2010}]%
        {DBLP:journals/igpl/ArtikisS10}
\bibfield{author}{\bibinfo{person}{Alexander Artikis} {and}
  \bibinfo{person}{Marek~J. Sergot}.} \bibinfo{year}{2010}\natexlab{}.
\newblock \showarticletitle{Executable specification of open multi-agent
  systems}.
\newblock \bibinfo{journal}{\emph{Logic Journal of the {IGPL}}}
  \bibinfo{volume}{18}, \bibinfo{number}{1} (\bibinfo{year}{2010}),
  \bibinfo{pages}{31--65}.
\newblock
\urldef\tempurl%
\url{https://doi.org/10.1093/jigpal/jzp071}
\showDOI{\tempurl}


\bibitem[\protect\citeauthoryear{Artikis, Sergot, and Paliouras}{Artikis
  et~al\mbox{.}}{2015}]%
        {DBLP:journals/tkde/ArtikisSP15}
\bibfield{author}{\bibinfo{person}{Alexander Artikis},
  \bibinfo{person}{Marek~J. Sergot}, {and} \bibinfo{person}{Georgios
  Paliouras}.} \bibinfo{year}{2015}\natexlab{}.
\newblock \showarticletitle{An Event Calculus for Event Recognition}.
\newblock \bibinfo{journal}{\emph{{IEEE} Trans. Knowl. Data Eng.}}
  \bibinfo{volume}{27}, \bibinfo{number}{4} (\bibinfo{year}{2015}),
  \bibinfo{pages}{895--908}.
\newblock
\urldef\tempurl%
\url{https://doi.org/10.1109/TKDE.2014.2356476}
\showDOI{\tempurl}


\bibitem[\protect\citeauthoryear{Artikis, Weidlich, Gal, Kalogeraki, and
  Gunopulos}{Artikis et~al\mbox{.}}{2013}]%
        {DBLP:conf/bigdataconf/ArtikisWGKG13}
\bibfield{author}{\bibinfo{person}{Alexander Artikis},
  \bibinfo{person}{Matthias Weidlich}, \bibinfo{person}{Avigdor Gal},
  \bibinfo{person}{Vana Kalogeraki}, {and} \bibinfo{person}{Dimitrios
  Gunopulos}.} \bibinfo{year}{2013}\natexlab{}.
\newblock \showarticletitle{Self-adaptive event recognition for intelligent
  transport management}. In \bibinfo{booktitle}{\emph{Proceedings of the {IEEE}
  International Conference on Big Data}}. \bibinfo{pages}{319--325}.
\newblock
\urldef\tempurl%
\url{https://doi.org/10.1109/BigData.2013.6691590}
\showDOI{\tempurl}


\bibitem[\protect\citeauthoryear{Beck, Dao{-}Tran, and Eiter}{Beck
  et~al\mbox{.}}{2018}]%
        {lars-rr17}
\bibfield{author}{\bibinfo{person}{Harald Beck}, \bibinfo{person}{Minh
  Dao{-}Tran}, {and} \bibinfo{person}{Thomas Eiter}.}
  \bibinfo{year}{2018}\natexlab{}.
\newblock \showarticletitle{{LARS:} {A} Logic-based framework for Analytic
  Reasoning over Streams}.
\newblock \bibinfo{journal}{\emph{Artif. Intell.}}  \bibinfo{volume}{261}
  (\bibinfo{year}{2018}), \bibinfo{pages}{16--70}.
\newblock
\urldef\tempurl%
\url{https://doi.org/10.1016/j.artint.2018.04.003}
\showDOI{\tempurl}


\bibitem[\protect\citeauthoryear{Cervesato and Montanari}{Cervesato and
  Montanari}{2000}]%
        {DBLP:conf/time/CervesatoM00}
\bibfield{author}{\bibinfo{person}{Iliano Cervesato} {and}
  \bibinfo{person}{Angelo Montanari}.} \bibinfo{year}{2000}\natexlab{}.
\newblock \showarticletitle{A Calculus of Macro-Events: Progress Report}. In
  \bibinfo{booktitle}{\emph{Seventh International Workshop on Temporal
  Representation and Reasoning, {TIME} 2000, Nova Scotia, Canada, July 7-9,
  2000}}. \bibinfo{pages}{47--58}.
\newblock
\urldef\tempurl%
\url{https://doi.org/10.1109/TIME.2000.856584}
\showDOI{\tempurl}


\bibitem[\protect\citeauthoryear{Chittaro and Montanari}{Chittaro and
  Montanari}{1996}]%
        {chittaro96}
\bibfield{author}{\bibinfo{person}{Luca Chittaro} {and} \bibinfo{person}{Angelo
  Montanari}.} \bibinfo{year}{1996}\natexlab{}.
\newblock \showarticletitle{Efficient Temporal Reasoning in the Cached Event
  Calculus}.
\newblock \bibinfo{journal}{\emph{Computational Intelligence}}
  \bibinfo{volume}{12} (\bibinfo{year}{1996}), \bibinfo{pages}{359--382}.
\newblock
\urldef\tempurl%
\url{https://doi.org/10.1111/j.1467-8640.1996.tb00267.x}
\showDOI{\tempurl}


\bibitem[\protect\citeauthoryear{Cugola and Margara}{Cugola and
  Margara}{2010}]%
        {DBLP:conf/debs/CugolaM10}
\bibfield{author}{\bibinfo{person}{Gianpaolo Cugola} {and}
  \bibinfo{person}{Alessandro Margara}.} \bibinfo{year}{2010}\natexlab{}.
\newblock \showarticletitle{{TESLA:} a formally defined event specification
  language}. In \bibinfo{booktitle}{\emph{Proceedings of the Fourth {ACM}
  International Conference on Distributed Event-Based Systems, {DEBS} 2010,
  Cambridge, United Kingdom, July 12-15, 2010}}. \bibinfo{pages}{50--61}.
\newblock
\urldef\tempurl%
\url{https://doi.org/10.1145/1827418.1827427}
\showDOI{\tempurl}


\bibitem[\protect\citeauthoryear{Cugola and Margara}{Cugola and
  Margara}{2012}]%
        {DBLP:journals/csur/CugolaM12}
\bibfield{author}{\bibinfo{person}{Gianpaolo Cugola} {and}
  \bibinfo{person}{Alessandro Margara}.} \bibinfo{year}{2012}\natexlab{}.
\newblock \showarticletitle{Processing flows of information: From data stream
  to complex event processing}.
\newblock \bibinfo{journal}{\emph{{ACM} Comput. Surv.}} \bibinfo{volume}{44},
  \bibinfo{number}{3} (\bibinfo{year}{2012}), \bibinfo{pages}{15:1--15:62}.
\newblock
\urldef\tempurl%
\url{https://doi.org/10.1145/2187671.2187677}
\showDOI{\tempurl}


\bibitem[\protect\citeauthoryear{Demers, Gehrke, Panda, Riedewald, Sharma, and
  White}{Demers et~al\mbox{.}}{2007}]%
        {DBLP:conf/cidr/DemersGPRSW07}
\bibfield{author}{\bibinfo{person}{Alan~J. Demers}, \bibinfo{person}{Johannes
  Gehrke}, \bibinfo{person}{Biswanath Panda}, \bibinfo{person}{Mirek
  Riedewald}, \bibinfo{person}{Varun Sharma}, {and} \bibinfo{person}{Walker~M.
  White}.} \bibinfo{year}{2007}\natexlab{}.
\newblock \showarticletitle{Cayuga: {A} General Purpose Event Monitoring
  System}. In \bibinfo{booktitle}{\emph{{CIDR} 2007, Third Biennial Conference
  on Innovative Data Systems Research, Asilomar, CA, USA, January 7-10, 2007,
  Online Proceedings}}. \bibinfo{pages}{412--422}.
\newblock
\urldef\tempurl%
\url{http://cidrdb.org/cidr2007/papers/cidr07p47.pdf}
\showURL{%
\tempurl}


\bibitem[\protect\citeauthoryear{Dong and Srivastava}{Dong and
  Srivastava}{2015}]%
        {DBLP:series/synthesis/2015Dong}
\bibfield{author}{\bibinfo{person}{Xin~Luna Dong} {and} \bibinfo{person}{Divesh
  Srivastava}.} \bibinfo{year}{2015}\natexlab{}.
\newblock \bibinfo{booktitle}{\emph{Big Data Integration}}.
\newblock \bibinfo{publisher}{Morgan {\&} Claypool Publishers}.
\newblock
\urldef\tempurl%
\url{https://doi.org/10.2200/S00578ED1V01Y201404DTM040}
\showDOI{\tempurl}


\bibitem[\protect\citeauthoryear{Dousson and Maigat}{Dousson and
  Maigat}{2007}]%
        {dousson07}
\bibfield{author}{\bibinfo{person}{Christophe Dousson} {and}
  \bibinfo{person}{Pierre~Le Maigat}.} \bibinfo{year}{2007}\natexlab{}.
\newblock \showarticletitle{Chronicle Recognition Improvement Using Temporal
  Focusing and Hierarchization}. In \bibinfo{booktitle}{\emph{{IJCAI} 2007,
  Proceedings of the 20th International Joint Conference on Artificial
  Intelligence, Hyderabad, India, January 6-12, 2007}}.
  \bibinfo{pages}{324--329}.
\newblock
\urldef\tempurl%
\url{http://ijcai.org/Proceedings/07/Papers/050.pdf}
\showURL{%
\tempurl}


\bibitem[\protect\citeauthoryear{Giatrakos, Alevizos, Artikis, Deligiannakis,
  and Garofalakis}{Giatrakos et~al\mbox{.}}{2019}]%
        {vldbj-cer-survey}
\bibfield{author}{\bibinfo{person}{Nikos Giatrakos}, \bibinfo{person}{Elias
  Alevizos}, \bibinfo{person}{Alexander Artikis}, \bibinfo{person}{Antonios
  Deligiannakis}, {and} \bibinfo{person}{Minos Garofalakis}.}
  \bibinfo{year}{2019}\natexlab{}.
\newblock \showarticletitle{Complex Event Recognition in the Big Data Era}.
\newblock \bibinfo{journal}{\emph{VLDB Journal}} (\bibinfo{year}{2019}).
\newblock


\bibitem[\protect\citeauthoryear{Grez, Riveros, and Ugarte}{Grez
  et~al\mbox{.}}{2019}]%
        {DBLP:conf/icdt/GrezRU19}
\bibfield{author}{\bibinfo{person}{Alejandro Grez}, \bibinfo{person}{Cristian
  Riveros}, {and} \bibinfo{person}{Mart{\'{\i}}n Ugarte}.}
  \bibinfo{year}{2019}\natexlab{}.
\newblock \showarticletitle{A Formal Framework for Complex Event Processing}.
  In \bibinfo{booktitle}{\emph{22nd International Conference on Database
  Theory, {ICDT} 2019, March 26-28, 2019, Lisbon, Portugal}}.
  \bibinfo{pages}{5:1--5:18}.
\newblock
\urldef\tempurl%
\url{https://doi.org/10.4230/LIPIcs.ICDT.2019.5}
\showDOI{\tempurl}


\bibitem[\protect\citeauthoryear{Jacox and Samet}{Jacox and Samet}{2007}]%
        {DBLP:journals/tods/JacoxS07}
\bibfield{author}{\bibinfo{person}{Edwin~H. Jacox} {and} \bibinfo{person}{Hanan
  Samet}.} \bibinfo{year}{2007}\natexlab{}.
\newblock \showarticletitle{Spatial join techniques}.
\newblock \bibinfo{journal}{\emph{{ACM} Trans. Database Syst.}}
  \bibinfo{volume}{32}, \bibinfo{number}{1} (\bibinfo{year}{2007}),
  \bibinfo{pages}{7}.
\newblock
\urldef\tempurl%
\url{https://doi.org/10.1145/1206049.1206056}
\showDOI{\tempurl}


\bibitem[\protect\citeauthoryear{Koutroumanis, Santipantakis, Glenis,
  Doulkeridis, and Vouros}{Koutroumanis et~al\mbox{.}}{2019}]%
        {DBLP:conf/edbt/KoutroumanisSGD19}
\bibfield{author}{\bibinfo{person}{Nikolaos Koutroumanis},
  \bibinfo{person}{Georgios~M. Santipantakis}, \bibinfo{person}{Apostolos
  Glenis}, \bibinfo{person}{Christos Doulkeridis}, {and}
  \bibinfo{person}{George~A. Vouros}.} \bibinfo{year}{2019}\natexlab{}.
\newblock \showarticletitle{Integration of Mobility Data with Weather
  Information}. In \bibinfo{booktitle}{\emph{Proceedings of the Workshops of
  the {EDBT/ICDT} 2019 Joint Conference, {EDBT/ICDT} 2019, Lisbon, Portugal,
  March 26, 2019.}}
\newblock
\urldef\tempurl%
\url{http://ceur-ws.org/Vol-2322/BMDA\_1.pdf}
\showURL{%
\tempurl}


\bibitem[\protect\citeauthoryear{Kowalski and Sergot}{Kowalski and
  Sergot}{1986}]%
        {DBLP:journals/ngc/KowalskiS86}
\bibfield{author}{\bibinfo{person}{Robert~A. Kowalski} {and}
  \bibinfo{person}{Marek~J. Sergot}.} \bibinfo{year}{1986}\natexlab{}.
\newblock \showarticletitle{A Logic-based Calculus of Events}.
\newblock \bibinfo{journal}{\emph{New Generation Comput.}} \bibinfo{volume}{4},
  \bibinfo{number}{1} (\bibinfo{year}{1986}), \bibinfo{pages}{67--95}.
\newblock
\urldef\tempurl%
\url{https://doi.org/10.1007/BF03037383}
\showDOI{\tempurl}


\bibitem[\protect\citeauthoryear{Liu, Rundensteiner, Greenfield, Gupta, Wang,
  Ari, and Mehta}{Liu et~al\mbox{.}}{2011}]%
        {DBLP:conf/sigmod/LiuRGGWAM11}
\bibfield{author}{\bibinfo{person}{Mo Liu}, \bibinfo{person}{Elke~A.
  Rundensteiner}, \bibinfo{person}{Kara Greenfield}, \bibinfo{person}{Chetan
  Gupta}, \bibinfo{person}{Song Wang}, \bibinfo{person}{Ismail Ari}, {and}
  \bibinfo{person}{Abhay Mehta}.} \bibinfo{year}{2011}\natexlab{}.
\newblock \showarticletitle{E-Cube: multi-dimensional event sequence analysis
  using hierarchical pattern query sharing}. In
  \bibinfo{booktitle}{\emph{Proceedings of the {ACM} {SIGMOD} International
  Conference on Management of Data, {SIGMOD} 2011, Athens, Greece, June 12-16,
  2011}}. \bibinfo{pages}{889--900}.
\newblock
\urldef\tempurl%
\url{https://doi.org/10.1145/1989323.1989416}
\showDOI{\tempurl}


\bibitem[\protect\citeauthoryear{Mei and Madden}{Mei and Madden}{2009}]%
        {DBLP:conf/sigmod/MeiM09}
\bibfield{author}{\bibinfo{person}{Yuan Mei} {and} \bibinfo{person}{Samuel
  Madden}.} \bibinfo{year}{2009}\natexlab{}.
\newblock \showarticletitle{ZStream: a cost-based query processor for
  adaptively detecting composite events}. In
  \bibinfo{booktitle}{\emph{Proceedings of the {ACM} {SIGMOD} International
  Conference on Management of Data, {SIGMOD} 2009, Providence, Rhode Island,
  USA, June 29 - July 2, 2009}}. \bibinfo{pages}{193--206}.
\newblock
\urldef\tempurl%
\url{https://doi.org/10.1145/1559845.1559867}
\showDOI{\tempurl}


\bibitem[\protect\citeauthoryear{Michelioudakis, Artikis, and
  Paliouras}{Michelioudakis et~al\mbox{.}}{2019}]%
        {DBLP:journals/ml/MichelioudakisA19}
\bibfield{author}{\bibinfo{person}{Evangelos Michelioudakis},
  \bibinfo{person}{Alexander Artikis}, {and} \bibinfo{person}{Georgios
  Paliouras}.} \bibinfo{year}{2019}\natexlab{}.
\newblock \showarticletitle{Semi-supervised online structure learning for
  composite event recognition}.
\newblock \bibinfo{journal}{\emph{Machine Learning}} \bibinfo{volume}{108},
  \bibinfo{number}{7} (\bibinfo{year}{2019}), \bibinfo{pages}{1085--1110}.
\newblock
\urldef\tempurl%
\url{https://doi.org/10.1007/s10994-019-05794-2}
\showDOI{\tempurl}


\bibitem[\protect\citeauthoryear{Miller and Shanahan}{Miller and
  Shanahan}{2002}]%
        {miller02}
\bibfield{author}{\bibinfo{person}{Rob Miller} {and} \bibinfo{person}{Murray
  Shanahan}.} \bibinfo{year}{2002}\natexlab{}.
\newblock \showarticletitle{Some Alternative Formulations of the Event
  Calculus}. In \bibinfo{booktitle}{\emph{Computational Logic: Logic
  Programming and Beyond, Essays in Honour of Robert A. Kowalski, Part {II}}}.
  \bibinfo{pages}{452--490}.
\newblock
\urldef\tempurl%
\url{https://doi.org/10.1007/3-540-45632-5\_17}
\showDOI{\tempurl}


\bibitem[\protect\citeauthoryear{Montali, Maggi, Chesani, Mello, and van~der
  Aalst}{Montali et~al\mbox{.}}{2013}]%
        {DBLP:journals/tist/MontaliMCMA13}
\bibfield{author}{\bibinfo{person}{Marco Montali},
  \bibinfo{person}{Fabrizio~Maria Maggi}, \bibinfo{person}{Federico Chesani},
  \bibinfo{person}{Paola Mello}, {and} \bibinfo{person}{Wil M.~P. van~der
  Aalst}.} \bibinfo{year}{2013}\natexlab{}.
\newblock \showarticletitle{Monitoring business constraints with the event
  calculus}.
\newblock \bibinfo{journal}{\emph{{ACM} {TIST}}} \bibinfo{volume}{5},
  \bibinfo{number}{1} (\bibinfo{year}{2013}), \bibinfo{pages}{17:1--17:30}.
\newblock
\urldef\tempurl%
\url{https://doi.org/10.1145/2542182.2542199}
\showDOI{\tempurl}


\bibitem[\protect\citeauthoryear{Paschke}{Paschke}{2006}]%
        {paschke05}
\bibfield{author}{\bibinfo{person}{Adrian Paschke}.}
  \bibinfo{year}{2006}\natexlab{}.
\newblock \showarticletitle{ECA-RuleML: An Approach combining {ECA} Rules with
  temporal interval-based {KR} Event/Action Logics and Transactional Update
  Logics}.
\newblock \bibinfo{journal}{\emph{CoRR}}  \bibinfo{volume}{abs/cs/0610167}
  (\bibinfo{year}{2006}).
\newblock
\showeprint[arxiv]{cs/0610167}
\urldef\tempurl%
\url{http://arxiv.org/abs/cs/0610167}
\showURL{%
\tempurl}


\bibitem[\protect\citeauthoryear{Paschke and Bichler}{Paschke and
  Bichler}{2008}]%
        {DBLP:journals/dss/PaschkeB08}
\bibfield{author}{\bibinfo{person}{Adrian Paschke} {and}
  \bibinfo{person}{Martin Bichler}.} \bibinfo{year}{2008}\natexlab{}.
\newblock \showarticletitle{Knowledge representation concepts for automated
  {SLA} management}.
\newblock \bibinfo{journal}{\emph{Decision Support Systems}}
  \bibinfo{volume}{46}, \bibinfo{number}{1} (\bibinfo{year}{2008}),
  \bibinfo{pages}{187--205}.
\newblock
\urldef\tempurl%
\url{https://doi.org/10.1016/j.dss.2008.06.008}
\showDOI{\tempurl}


\bibitem[\protect\citeauthoryear{Patroumpas, Alevizos, Artikis, Vodas, Pelekis,
  and Theodoridis}{Patroumpas et~al\mbox{.}}{2017}]%
        {DBLP:journals/geoinformatica/PatroumpasAAVPT17}
\bibfield{author}{\bibinfo{person}{Kostas Patroumpas}, \bibinfo{person}{Elias
  Alevizos}, \bibinfo{person}{Alexander Artikis}, \bibinfo{person}{Marios
  Vodas}, \bibinfo{person}{Nikos Pelekis}, {and} \bibinfo{person}{Yannis
  Theodoridis}.} \bibinfo{year}{2017}\natexlab{}.
\newblock \showarticletitle{Online event recognition from moving vessel
  trajectories}.
\newblock \bibinfo{journal}{\emph{GeoInformatica}} \bibinfo{volume}{21},
  \bibinfo{number}{2} (\bibinfo{year}{2017}), \bibinfo{pages}{389--427}.
\newblock
\urldef\tempurl%
\url{https://doi.org/10.1007/s10707-016-0266-x}
\showDOI{\tempurl}


\bibitem[\protect\citeauthoryear{Schultz{-}M{\o}ller, Migliavacca, and
  Pietzuch}{Schultz{-}M{\o}ller et~al\mbox{.}}{2009}]%
        {DBLP:conf/debs/Schultz-MollerMP09}
\bibfield{author}{\bibinfo{person}{Nicholas~Poul Schultz{-}M{\o}ller},
  \bibinfo{person}{Matteo Migliavacca}, {and} \bibinfo{person}{Peter~R.
  Pietzuch}.} \bibinfo{year}{2009}\natexlab{}.
\newblock \showarticletitle{Distributed complex event processing with query
  rewriting}. In \bibinfo{booktitle}{\emph{Proceedings of the Third {ACM}
  International Conference on Distributed Event-Based Systems, {DEBS} 2009,
  Nashville, Tennessee, USA, July 6-9, 2009}}.
\newblock
\urldef\tempurl%
\url{https://doi.org/10.1145/1619258.1619264}
\showDOI{\tempurl}


\bibitem[\protect\citeauthoryear{Tsilionis, Artikis, and Paliouras}{Tsilionis
  et~al\mbox{.}}{2019}]%
        {eftsilio_DEBS}
\bibfield{author}{\bibinfo{person}{Efthimis Tsilionis},
  \bibinfo{person}{Alexander Artikis}, {and} \bibinfo{person}{Georgios
  Paliouras}.} \bibinfo{year}{2019}\natexlab{}.
\newblock \showarticletitle{Incremental Event Calculus for Run-Time Reasoning}.
  In \bibinfo{booktitle}{\emph{Proceedings of the 13th {ACM} International
  Conference on Distributed and Event-based Systems, {DEBS} 2019, Darmstadt,
  Germany, June 24-28, 2019.}} \bibinfo{pages}{79--90}.
\newblock
\urldef\tempurl%
\url{https://doi.org/10.1145/3328905.3329504}
\showDOI{\tempurl}


\bibitem[\protect\citeauthoryear{Vlassopoulos and Artikis}{Vlassopoulos and
  Artikis}{2017}]%
        {DBLP:conf/commonsense/VlassopoulosA17}
\bibfield{author}{\bibinfo{person}{Christos Vlassopoulos} {and}
  \bibinfo{person}{Alexander Artikis}.} \bibinfo{year}{2017}\natexlab{}.
\newblock \showarticletitle{Towards {A} Simple Event Calculus for Run-Time
  Reasoning}. In \bibinfo{booktitle}{\emph{Proceedings of the Thirteenth
  International Symposium on Commonsense Reasoning, {COMMONSENSE} 2017, London,
  UK, November 6-8, 2017.}}
\newblock
\urldef\tempurl%
\url{http://ceur-ws.org/Vol-2052/paper20.pdf}
\showURL{%
\tempurl}


\bibitem[\protect\citeauthoryear{Zhang, Diao, and Immerman}{Zhang
  et~al\mbox{.}}{2014}]%
        {sasep}
\bibfield{author}{\bibinfo{person}{Haopeng Zhang}, \bibinfo{person}{Yanlei
  Diao}, {and} \bibinfo{person}{Neil Immerman}.}
  \bibinfo{year}{2014}\natexlab{}.
\newblock \showarticletitle{On complexity and optimization of expensive queries
  in complex event processing}. In \bibinfo{booktitle}{\emph{International
  Conference on Management of Data, {SIGMOD} 2014, Snowbird, UT, USA, June
  22-27, 2014}}. \bibinfo{pages}{217--228}.
\newblock
\urldef\tempurl%
\url{https://doi.org/10.1145/2588555.2593671}
\showDOI{\tempurl}


\end{thebibliography}

\end{document}